\documentclass[moor,nonblindrev]{informs1} 

\usepackage{natbib}
 \NatBibNumeric
 \def\bibfont{\small}%
 \bibpunct[, ]{[}{]}{,}{n}{}{,}%
 
\usepackage{algorithm}
\usepackage{algpseudocode}
\usepackage{enumitem}
\usepackage{wasysym}
\usepackage{subfigure}

 \usepackage[colorlinks=true,breaklinks=true,bookmarks=false,urlcolor=blue, citecolor=blue,linkcolor=blue,bookmarksopen=false,draft=false]{hyperref}

\def\EMAIL#1{\href{mailto:#1}{#1}}
\def\URL#1{\href{#1}{#1}}         

\TheoremsNumberedThrough     

\EquationsNumberedThrough    

\newlength \myindent
\begin{document}


\RUNAUTHOR{Wen and Van Roy}

\RUNTITLE{Efficient Reinforcement Learning in Deterministic Systems}


\newcommand{\state}{\mathcal{S}}

\newcommand{\action}{\mathcal{A}}

\newcommand{\mdp}{\mathcal{M}}
\newcommand{\mdpset}{\mathbb{M}}

\newcommand{\SAT}{\mathcal{Z}}
\newcommand{\subSAT}{\tilde{\mathcal{Z}}}

\newcommand{\sat}{z}

\newcommand{\Fspace}{\mathcal{F}}

\newcommand{\Fsubspace}{\tilde{\mathcal{F}}}

\newcommand{\Qspace}{\mathcal{Q}}

\newcommand{\dimM}[1]{\mathrm{dim}_{\mathrm{E}}[ #1 ]}
\newcommand{\spann}{\mathrm{span}}

\newcommand{\cat}{\stackrel{\frown}{}}

\newcommand{\BE}{\begin{eqnarray}}
\newcommand{\EE}{\end{eqnarray}}
\newcommand{\reg}{\mathrm{Regret}}
\newcommand{\indicator}{\mathbf{1}}
\newcommand{\rank}{\mathrm{rank}}

\newcommand{\Qopt}{Q^{\uparrow}}
\newcommand{\Qpes}{Q^{\downarrow}}
\newcommand{\NULL}{\mathrm{NULL}}
\newcommand{\Rmax}{\overline{R}}

\newcommand{\nn}{\nonumber \\}

\newcommand{\RL}{\tilde{\mu}}

\newcommand{\BO}[1]{O \left( #1 \right)}

\newcommand{\CC}{\mathcal{C}}

\TITLE{Efficient Reinforcement Learning in Deterministic Systems with Value Function Generalization}


\ARTICLEAUTHORS{%
\AUTHOR{Zheng Wen}
\AFF{Adobe Research, \EMAIL{zwen@adobe.com}, \URL{}}
\AUTHOR{Benjamin Van Roy}
\AFF{Stanford University, \EMAIL{bvr@stanford.edu}, \URL{}}
} 

\ABSTRACT{%
We consider the problem of reinforcement learning over episodes of a finite-horizon deterministic system 
and as a solution propose {\it optimistic constraint propagation (OCP)}, an algorithm
designed to synthesize efficient exploration and value function generalization.  We establish that when
the true value function $Q^*$ lies within a known hypothesis class $\mathcal{Q}$, OCP selects optimal actions
over all but at most $\dimM{\mathcal{Q}}$ episodes, where ${\rm dim_E}$ denotes the
{\it eluder dimension}.  We establish
further efficiency and asymptotic performance guarantees that apply even if $Q^*$ does not lie in $\mathcal{Q}$, for 
the special case where $\mathcal{Q}$ is the span of pre-specified indicator functions over disjoint sets.
We also discuss the computational complexity of OCP 
and present computational results involving two illustrative examples.
%
}%



\maketitle

\section{Introduction}
A growing body of work on efficient reinforcement learning provides algorithms with guarantees on sample and 
computational efficiency (see, e.g., \citep{Kearns2002,Brafman2002,Auer2006,Strehl2006,Bartlett2009,Jaksch2010} and references therein).  
This literature highlights the point that an effective exploration scheme is critical to the design of any efficient reinforcement learning algorithm.
In particular, popular exploration schemes such as $\epsilon$-greedy, Boltzmann, and knowledge gradient (see \citep{ryzhov2010approximate}) can require 
learning times that grow exponentially in the number of states and/or the planning horizon (see \citep{whitehead2014complexity, strehl2007probably}).

The aforementioned literature focusses on {\it tabula rasa} learning; that is, algorithms aim to learn with
little or no prior knowledge about transition probabilities and rewards.  Such algorithms require learning 
times that grow at least linearly with the number of states.  Despite the valuable insights
that have been generated through their design and analysis, these algorithms are of limited practical 
import because state spaces in most contexts of practical interest are enormous.  There is a need for 
algorithms that generalize from past experience in order to learn how to make effective decisions in reasonable time.

There has been much work on reinforcement learning algorithms that generalize 
(see, e.g., \citep{Bertsekas1996,Sutton1998,2010Szepesvari,Powell2011} and references therein).
Most of these algorithms do not come with statistical or computational efficiency guarantees, though there are
a few noteworthy exceptions, which we now discuss.
A number of results treat policy-based algorithms (see \citep{Kakade2003,Azar2013} and references therein), in which 
the goal is to select high-performers among a pre-specified collection of policies as learning progresses.
Though interesting results have been produced in this line of work, each entails quite restrictive assumptions 
or does not make strong guarantees.
Another body of work focuses on model-based algorithms.  An algorithm proposed by \citet{Kearns1999} 
fits a factored model to observed data and makes decisions based on the fitted model.  The authors establish a sample complexity bound
that is polynomial in the number of model parameters rather than the number of states, but the algorithm is computationally
intractable because of the difficulty of solving factored MDPs.  
\citet{Lattimore2013} propose a novel algorithm for the case
where the true environment is known to belong to a finite or compact class of models, and shows that its sample complexity is polynomial
in the cardinality of the model class if the model class is finite, or the $\epsilon$-covering-number if the model class is compact. 
Though this result is theoretically interesting, for most model classes of interest, the $\epsilon$-covering-number is enormous since it 
typically grows exponentially in the number of free parameters.
\citet{Ortner2012} establish a regret bound for an algorithm that applies to problems with
continuous state spaces and H{\"o}lder-continuous rewards and transition kernels.  Though the results
represent an interesting contribution to the literature, a couple of features of the regret bound weaken its practical implications.
First, regret grows linearly with the H{\"o}lder constant of the transition kernel, which for most contexts of practical relevance
grows exponentially in the number of state variables.  Second, the dependence on time becomes arbitrarily close to linear as
the dimension of the state space grows.
\citet{pazis2013pac} also consider problems with
continuous state spaces.  They assume that the Q-functions are Lipschitz-continuous or H{\"o}lder-continuous
and establish a sample complexity bound.
Though the results are interesting and significant, the sample complexity bound is log-linear in the covering number of the 
state-action space, which also typically grows exponentially in the number of free parameters for most practical problems.
Reinforcement learning in linear systems with quadratic cost
is treated in \citet{Abbasi-Yadkori2011}.  The method proposed is shown to realize regret that grows with the square root of time.
The result is interesting and the property is desirable, but to the best of our knowledge, expressions derived for regret in the analysis 
exhibit an exponential dependence on the number of state variables, and further, we are not aware of a computationally efficient 
way of implementing the proposed method.  This work was extended by \citet{Ibrahimi2012} to address linear systems with sparse structure.
Here, there are efficiency guarantees that scale gracefully with the number of state variables, but only under sparsity
and other technical assumptions.

The most popular approach to generalization in the applied reinforcement learning literature involves fitting parameterized value 
functions.  Such approaches
relate closely to supervised learning in that they learn functions from state-action pairs to value, though a difference is that
value is influenced by action and observed only through delayed feedback.  One advantage over model learning approaches
is that, given a fitted value function, decisions can be made without solving an often intractable control problem.
We see this as a promising direction, though there currently is a lack of theoretical results that provide attractive bounds on
learning time with value function generalization.  
A relevant paper along these lines is \citep{Li2010}, which 
studies efficient reinforcement learning with value function generalization in the 
KWIK framework (see \citep{li2011knows}) and reduces the problem
to efficient KWIK online regression.
However, the authors do not show how to solve the general KWIK online regression
problem efficiently, and it is not even clear whether this is possible.
Thus, though the result of \citet{Li2010} is interesting, it does not 
provide a provably efficient algorithm for general reinforcement learning problems.
However, it is worth mentioning that \citet{li2011knows} has provided a solution to KWIK online regression with 
deterministic linear functions. As we will discuss later, this can be seen as a special case of the coherent learning problems we consider in Section \ref{sec:coherent_hypothesis_class}.

An important challenge that remains is to couple exploration and value function generalization in a provably effective way, and 
in particular, to establish sample and computational efficiency guarantees that scale gracefully with the planning horizon and model complexity.
In this paper, we aim to make progress in this direction.  To start with a simple context, we restrict our attention to
deterministic systems that evolve over finite time horizons, and we consider episodic learning, in which
an agent repeatedly interacts with the same system.  As a solution to the problem, we 
propose {\it optimistic constraint propagation (OCP)}, a computationally efficient reinforcement learning algorithm
designed to synthesize efficient exploration and value function generalization.  We establish that when
the true value function $Q^*$ lies within the hypothesis class $\Qspace$, OCP selects optimal actions
over all but at most $\dimM{\mathcal{Q}}$ episodes.  Here, ${\rm dim_E}$ denotes the
{\it eluder dimension}, which quantifies complexity of the hypothesis class.  A corollary of this result
is that regret is bounded by a function that is constant over time and linear in the problem horizon and 
eluder dimension.

To put our aforementioned result in perspective, it is useful to relate it to other lines of work.  Consider first the broad area
of reinforcement learning algorithms that fit value functions, such as SARSA \citep{Rummery1994}.
Even with the most commonly used sort of hypothesis class $\Qspace$, which is made up of
linear combinations of fixed basis functions, and even when the hypothesis class contains the true value function $Q^*$, 
there are no guarantees that these algorithms will efficiently learn to make near-optimal decisions.
On the other hand, our result implies that OCP attains near-optimal performance in time that scales linearly with
the number of basis functions.
Now consider the more specialized context of a deterministic linear system with quadratic cost and a finite time horizon.
The analysis of \citet{Abbasi-Yadkori2011} can be leveraged to produce regret bounds that scale exponentially in the 
number of state variables.  On the other hand, using a hypothesis space $\Qspace$ consisting of quadratic functions
of state-action pairs, the results of this paper show that OCP behaves near optimally within time that scales
quadratically in the number of state and action variables.

We also establish efficiency and asymptotic performance guarantees that apply to agnostic reinforcement learning, where 
$Q^*$ does not necessarily lie in $\mathcal{Q}$.  In particular, we consider the case where $\mathcal{Q}$ is the span of 
pre-specified indicator functions over disjoint sets.  Our results here add to the literature on agnostic reinforcement learning 
with such a hypothesis class \citep{Singh1994,TsitsiklisVR1996,Gordon1995,VanRoy2006}.  Prior work in this
area has produced interesting algorithms and insights, as well as bounds on performance loss associated with potential limits
of convergence, but no convergence or efficiency guarantees. These results build on and add to those reported in an earlier paper that we published in proceedings of a conference \cite{wen2013efficient}.

In addition to establishing theoretical results,
we present computational results involving two illustrative examples: 
a synthetic deterministic Markov chain and the inverted pendulum control problem considered in \citet{lagoudakis2002least}.
We compare OCP against least-squares value iteration (LSVI), a classical reinforcement learning algorithm.
In both experiments, the performance of OCP is orders of magnitude better than that of LSVI.
It is worth mentioning that in the inverted pendulum example, we consider a case in which there are small stochastic disturbances additive to the control. This result shows that, though OCP is designed for deterministic systems, it might also work well in stochastic environments, especially when the magnitude of the stochastic disturbances is small.

Finally, it is worth pointing out that reinforcement learning algorithms are often used to approximate solutions to large-scale dynamic programs, 
where the system models are \emph{known}.  By \emph{known}, we mean that, given sufficient compute power, one can determine the expected 
single-period rewards and transition probabilities with any desired level of accuracy in the absence of any additional empirical data.
In such contexts, there is no need for statistical learning as challenges are purely computational. Nevertheless, reinforcement learning algorithms make up popular 
solution techniques for such problems, and our algorithm and results also serve as contributions to the field of approximate dynamic programming.
Specifically, prior approximate dynamic programming algorithms that fit a linear combination of basis functions to the value function, even when the optimal 
value function is within the span, come with no guarantees that a near-optimal policy can be computed efficiently.  In this paper, we establish such a guarantee for OCP.

\section{Episodic Reinforcement Learning in Deterministic Systems}\label{sec:problem_formulation}

We consider a class of reinforcement learning problems in which an agent repeatedly interacts with
an unknown discrete-time deterministic finite-horizon Markov decision process (MDP). 
Each interaction is referred to as an \emph{episode}, and the agent's objective is to maximize the expected cumulative reward over episodes.
The system is identified by a sextuple $\mdp=\left(\state, \action, H, F, R, S\right)$, where 
$\state$ is the state space,
$\action$ is the action space, $H$ is the horizon,
$F$ is a system function, $R$ is a reward function and
$S$ is a sequence of states.
If action $a \in \action$ is selected while the system is in state $x \in \state$ at period $t=0,1,\cdots, H-1$, a reward
of $R_t(x,a)$ is realized; furthermore, if $t<H-1$, the 
state transitions to $F_t(x,a)$.
Each episode terminates at period $H-1$, and then a new episode begins.
The initial state of episode $j$ is the $j$th element of $S$.

To represent the history of actions and observations over multiple episodes, we will often index variables
by both episode and period.  For example, 
$x_{j,t}$ and $a_{j,t}$ denote the state and action at period $t$ of episode $j$,
where $j=0,1,\cdots$ and $t=0,1,\cdots, H-1$. 
To count the total number of steps since the agent started learning,
we say period $t$ in episode $j$ is time $jH+t$.

A (deterministic) policy $\mu = (\mu_0,\ldots,\mu_{H-1})$ is a sequence of functions, each mapping $\state$ to $\action$.
For each policy $\mu$, define a value function $V^\mu_t(x) = \sum_{\tau=t}^{H-1} R_{\tau} (x_\tau,a_\tau)$, where $x_t = x$,
$x_{\tau+1} = F_{\tau} (x_\tau,a_\tau)$, and $a_\tau = \mu_\tau(x_\tau)$.  The optimal value function is defined by
$V^*_t(x) = \sup_\mu V^\mu_t(x)$.  A policy $\mu^*$ is said to be optimal if $V^{\mu^*} = V^*$.  Throughout this
paper, we will restrict attention to systems $\mdp=\left(\state, \action, H, F, R, S\right)$ that admit optimal policies.
Note that this restriction incurs no loss of generality when the action space is finite.

It is also useful to define an action-contingent optimal value function: 
$Q^*_t(x,a) = R_t(x,a)+V^*_{t+1}(F_t(x,a))$ for $t < H-1$, and 
$Q^*_{H-1}(x,a) = R_{H-1}(x,a)$.
Then, a policy $\mu^*$ is optimal if $\mu_t^*(x) \in \argmax_{a \in \action} Q^*_t(x,a)$ for all $(x,t)$.

This paper considers a reinforcement learning framework in which the agent initially knows the state space
$\state$, the action space $\action$, the horizon $H$, and possibly some prior information about the value function,
but does not know anything else about the system function $F$, the reward function $R$, or the sequence of the initial states $S$.
A reinforcement learning algorithm generates each action $a_{j,t}$ based on observations made up
to the $t$th period of the $j$th episode, including all states, actions, and rewards observed in previous episodes and
earlier in the current episode, as well as $\state$, $\action$, $H$, and 
possible prior information.  In each episode, the algorithm realizes reward
$R^{(j)} = \sum_{t=0}^{H-1} R_t \left(x_{j,t}, a_{j,t} \right)$.
Note that $R^{(j)} \le V^*_0(x_{j,0})$ for each $j$th episode.
To quantify the performance of a reinforcement learning algorithm, for any $\epsilon \geq 0$, we define the 
\emph{$\epsilon$-suboptimal sample complexity} of that algorithm as the number of episodes $J_L$ for which $R^{(j)} < V^*_0(x_{j,0})- \epsilon$. 
Moreover, we say a reinforcement 
learning algorithm is sample efficient in a given setting if for some reasonable choice of $\epsilon$, 
the worst-case $\epsilon$-suboptimal sample complexity of that algorithm is small for that setting.
Note that if the reward function $R$ is bounded, with 
$|R_t (x,a)| \leq \overline{R}$ for all $(x,a,t)$, then a bound on $\epsilon$-suboptimal sample complexity
$J_L$ also implies a bound on regret 
over episodes
experienced prior to time $T$,
defined by
$\text{Regret} (T) = \sum_{j=0}^{\lfloor T / H\rfloor-1} (V^*_0(x_{j,0}) - R^{(j)})$.
In particular,
$\text{Regret}(T) \leq 2 \overline{R} H J_L + \epsilon \lfloor T / H\rfloor $.

\section{Inefficient Exploration Schemes}
\label{sec:inefficient}
Before proceeding, it is worth pointing out that for the reinforcement learning problem proposed above, a number of popular exploration schemes give rise to exponentially large
regret.  Even in the {\it tabula rasa} case, Boltzmann\footnote{Notice
that in this paper, we assume that the state transition model of the deterministic system is unknown.
Some literature (see \citep{neu2010online} and references therein) consider settings in which the state transition model is known but the reward function is unknown,
and establish that exploration schemes similar to Boltzmann exploration
achieve regret polynomial in $H$ (or, more generally, a notion of mixing time) and $|\state|$.} 
and $\epsilon$-greedy exploration schemes (see, e.g., \citep{Powell2007ADP}), for example, lead to worst-case regret exponential in $H$ and/or $|\state|$.  Also,  
the knowledge gradient exploration scheme (see, e.g., \citep{Powell2011} and \citep{ryzhov2010approximate}) can converge to suboptimal policies, and even when the ultimate policy is optimal, 
regret can grow exponentially in $H$ and/or $|\state|$.  Thus, even for the {\it tabula rasa} case, efficient exploration schemes are necessary
for an algorithm to achieve regret polynomial in $H$ and $|\state|$.

\begin{figure}[h]
\centering
\includegraphics[scale=0.55]{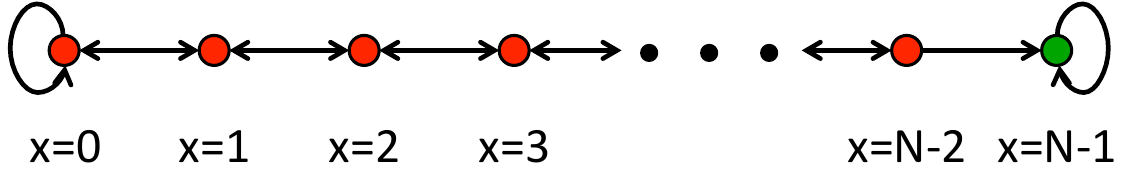}
\caption{Deterministic system for which Boltzmann and $\epsilon$-greedy exploration are inefficient.}
\label{fig:example1}
\end{figure}

To illustrate how simple exploration schemes give rise to exponentially large regret and how OCP will mitigate that, consider the following simple example.
\begin{example}
\label{example:chain}
Consider the deterministic system illustrated in Figure \ref{fig:example1}.  Each node represents a state, and each arrow 
corresponds to a possible state transition. The state space is
$\state=\left \{ 0,1,\cdots, N-1 \right \}$ and the action space is $\action=\left \{a^{(1)},a^{(2)} \right \}$.
If the agent takes action $a^{(1)}$ at state $x=0,1,\cdots, N-2$,
the state transitions to
$y=[x-1]^+$.  On the other hand, if the agent takes action $a^{(2)}$ at state $x=0,1,\cdots, N-2$, the state
transitions to $y=x+1$. State $N-1$ 
is absorbing.
We assume a reward of $0$ is realized upon any transition from node $ 0,1,\cdots, N-2$ 
and a reward of $1$ is realized upon any transition 
from node $N-1$.  We take the horizon $H$ to be equal to the number of states $N$.  The initial state
in any episode is $0$. 
\end{example}

For the example we have described, the only way to realize any reward in an episode is to select action $a^{(2)}$ over $N-1$ consecutive time periods.
Starting with no special knowledge about the system and with default estimates of $0$ for each period-state-action value $Q_t^*(x,a)$, 
Boltzmann and $\epsilon$-greedy can only discover the reward opportunity via random wandering,
which requires $2^{|\state|-1}$ episodes in expectation.  This translates to a lower bound on expected regret\footnote{Since Boltzmann exploration and $\epsilon$-greedy exploration are randomized exploration schemes, we should measure the performance of LSVI with Boltzmann/$\epsilon$-greedy exploration with expected regret. We use the same symbol $\reg(T)$ for the expected regret since the regret defined in this paper can be viewed as a special case of the expected regret.}:
\begin{equation}
\label{eqn:example_1_regret_lb_1}
\reg (T) \geq  \left(2^{|\state|-1} - 1 \right) \left(1 - \left[1 - 2^{-(|\state|-1)}\right]^{\lfloor T/H \rfloor} \right),
\end{equation}
which implies that 
\begin{equation}
\label{eqn:example_1_regret_lb_2}
\liminf_{T \rightarrow \infty} \textrm{Regret} (T) \geq 2^{|\state|-1} - 1.
\end{equation}

One way to dramatically reduce regret is through optimism.  In particular, if a learning agent begins with an initial estimate of $1$ for each period-state-action 
value $Q_t^*(x,a)$, this incentivizes selection of actions not yet tried and can reduce the dependence of regret on $|\mathcal{S}|$ to linear in the {\it tabula rasa} case.

The situation becomes more complex, however, when the agent generalizes across period, states, and/or actions.  
Generalization means altering a value estimate at one period-state-action triple based on observations made at others.  
An incorrect generalization can turn an optimistic estimate into a pessimistic one.  OCP is an algorithm that 
generalizes in a manner that prevents this from happening.  As we will establish, by retaining optimism, OCP guarantees
low regret.

\section{Optimistic Constraint Propagation}\label{sec:algorithm}

Our reinforcement learning algorithm -- optimistic constraint propagation (OCP) -- takes as input the state space $\state$,
the action space $\action$, the horizon $H$, and a hypothesis class $\Qspace$ of candidates for $Q^*$.  The algorithm maintains
a sequence of subsets of $\Qspace$ and a sequence of scalar ``upper bounds'', which summarize constraints that past 
experience suggests for ruling out hypotheses.  Each constraint in this sequence is specified by a state $x \in \state$, an action $a \in \action$,
a period $t = 0,\ldots,H-1$, and an interval $[L,U] \subseteq \Re$, and takes the form $\{Q \in \Qspace : L \leq Q_t(x,a) \leq U\}$.  
The upper bound of the constraint is $U$.
Given a sequence $\CC = (\CC_1, \ldots, \CC_{|\CC|})$ of such constraints
and upper bounds $\mathcal{U} = (U_1, \ldots, U_{|\CC|})$, for any $i,j = 1, \cdots, |\CC|$
s.t. $i \neq j$, we say $\CC_i < \CC_j$, or constraint $\CC_i$ has higher priority than $\CC_j$, if
(1) $U_i < U_j$ or (2) $U_i = U_j$ and $j>i$. That is, priority is assigned first based on upper bound, with smaller upper bound preferred,
and then, in the event of ties in upper bound, based on position in the sequence, with more recent experience (larger index) preferred.
A set $\Qspace_{\CC}$ is defined constructively by Algorithm \ref{alg:ConstraintSelection}. 
Note that
if the constraints do not conflict then $\Qspace_{\CC} = \CC_1\cap \cdots \cap \CC_{|\CC|}$.

\begin{algorithm}[th]
\caption{Constraint Selection}
  \label{alg:ConstraintSelection}
\begin{algorithmic}
\Require $\Qspace$, $\CC$
\State $\Qspace_{\CC} \leftarrow \Qspace$
\State Sort constraints in $\CC$ s.t. $\CC_{k_1} < \CC_{k_2} < \cdots < \CC_{k_{|\CC|}}$
\For{$\tau=1$ to $|\CC|$}
\If{$\Qspace_{\mathcal{C}} \cap \mathcal{C}_{k_\tau} \neq \varnothing$}
\State $\Qspace_{\mathcal{C}} \leftarrow \Qspace_{\mathcal{C}} \cap \mathcal{C}_{k_\tau}$
\EndIf
\EndFor
\State \Return $\Qspace_{\mathcal{C}}$
\end{algorithmic}
\end{algorithm}



OCP, presented below as Algorithm \ref{alg:OCP}, at each time $t$ computes for the current state $x_{j,t}$ and each action $a$ the greatest 
state-action value $Q_t(x_{j,t},a)$ among functions in $\Qspace_{\mathcal{C}}$ and selects an action that attains the maximum.  In other words,
an action is chosen based on the most optimistic feasible outcome subject to constraints.  The subsequent reward and
state transition give rise to a new constraint that is used to update ${\mathcal{C}}$.  
Note that the update of $\mathcal{C}$ is postponed until one episode is completed.

\begin{algorithm}[th]
\caption{Optimistic Contraint Propagation}
  \label{alg:OCP}
\begin{algorithmic}
\Require $\state$, $\action$, $H$, $\Qspace$ 
\State Initialize $\mathcal{C} \leftarrow \varnothing$
\For{episode $j=0,1,\cdots$}
\State Set $\mathcal{C}' \leftarrow \mathcal{C}$
\For{period $t=0,1,\cdots, H-1$}
\State Apply $a_{j,t} \in \argmax_{a \in \action} \sup_{Q \in \Qspace_{\mathcal{C}}} Q_t(x_{j,t}, a)$
\If{$t<H-1$}
\State $U_{j,t} \leftarrow \sup_{Q \in \Qspace_{\mathcal{C}}} \left(R_t (x_{j,t}, a_{j,t}) + \sup_{a \in \action} Q_{t+1} \left( x_{j,t+1}, a \right)\right)$
\State $L_{j,t}  \leftarrow \inf_{Q \in \Qspace_{\mathcal{C}}} \left(R_t (x_{j,t}, a_{j,t}) +  \sup_{a \in \action} Q_{t+1} \left( x_{j,t+1}, a \right)\right)$
\Else
\State $U_{j,t}  \leftarrow R_t (x_{j,t}, a_{j,t})$, $L_{j,t} \leftarrow R_t (x_{j,t}, a_{j,t})$
\EndIf
\State $\mathcal{C}' \leftarrow  \mathcal{C}' \cat \left \lbrace Q \in \Qspace:\, L_{j,t} \le Q_t (x_{j,t}, a_{j,t}) \le U_{j,t} \right \rbrace$
\EndFor
\State Update $\mathcal{C} \leftarrow \mathcal{C}'$
\EndFor
\end{algorithmic}
\end{algorithm}

As we will prove in Lemma \ref{lemma:tech0}, if $Q^* \in \Qspace$ then
each constraint appended to $\mathcal{C}$ does not rule out $Q^*$, and therefore, the sequence of 
sets $\Qspace_{\mathcal{C}}$ generated as the algorithm progresses
is decreasing and contains $Q^*$ in its intersection.  In the agnostic case,
where $Q^*$ may not lie in $\Qspace$, new constraints can be inconsistent with previous constraints, in which case 
selected previous constraints are relaxed as determined by Algorithm \ref{alg:ConstraintSelection}.

Let us briefly discuss several contexts of practical relevance and/or theoretical interest in which OCP can be applied.
\begin{itemize}
\item \textbf{Finite state/action tabula rasa case.}  With finite state and action spaces, $Q^*$ can be represented
as a vector, and without special prior knowledge, it is natural to let $\Qspace = \Re^{|\state| \cdot |\action| \cdot H}$.
\item \textbf{Polytopic prior constraints.}  Consider the aforementioned example, but suppose that we have prior knowledge that 
$Q^*$ lies in a particular polytope.  Then we can let $\Qspace$ be that polytope and again apply OCP.
\item \textbf{Linear systems with quadratic cost (LQ).}  In this classical control model, if $\state = \Re^n$, $\action = \Re^m$, 
and $R$ is a positive semidefinite quadratic, then for each $t$, $Q^*_t$ is known to be a positive semidefinite quadratic, and it is natural
to let $\Qspace = \Qspace_0^H$ with $\Qspace_0$ denoting the set of positive semidefinite quadratics.
\item \textbf{Finite hypothesis class.} Consider a context when we have prior knowledge that $Q^*$ can be well approximated by
some element in a finite hypothesis class. Then we can let $\Qspace$ be that finite hypothesis class and apply OCP. 
This scenario is of particular interest from the
perspective of learning theory. 
Note that this context entails agnostic learning, which is accommodated by OCP.
\item \textbf{Linear combination of features.} It is often effective to hand-select a set of features $\phi_1,\ldots,\phi_K$, 
each mapping $\state \times \action$ to $\Re$, and, then for each $t$, aiming to compute weights $\theta^{(t)} \in \Re^K$ so that
$\sum_k \theta^{(t)}_k \phi_k$ approximates $Q^*_t$ without knowing for sure that $Q^*_t$ lies in the span of the features.
To apply OCP here, we would let $\Qspace = \Qspace_0^H$ with $\Qspace_0 = \text{span}(\phi_1,\ldots,\phi_K)$.
Note that this context also entails agnostic learning.
\item \textbf{State aggregation.} This is a special case of the ``linear combination of features" case discussed above.
Specifically, for any $t=0,1,\cdots,H-1$,
the state-action space at period $t$,
$\SAT_t= \left \lbrace (x,a, t): \, x \in \state, a \in \action \right \rbrace$,
is partitioned into $K_t$ disjoint subsets $\SAT_{t,1}, \SAT_{t,2}, \cdots, \SAT_{t, K_t}$,
and we choose as features indicator functions for partition $\SAT_{t,k}$'s.
\item \textbf{Sigmoid.} If it is known that rewards are only received upon transitioning to the terminal state and take values between
$0$ and $1$, it might be appropriate to use a variation of the aforementioned feature based model that applies a sigmoidal function to the linear
combination.
In particular, we could have $\Qspace = \Qspace_0^H$ with $\Qspace_0 = \left\{\psi\left(\sum_k \theta_k \phi_k(\cdot)\right): \theta \in \Re^K\right\}$,
where $\psi(z) = e^z / (1+e^z)$.
\item \textbf{Sparse linear combination of features.} 
Another case of potential interest is where $Q^*$ can be encoded by a 
sparse linear combination of a large number of features $\phi_0, \cdots, \phi_K$. In particular, suppose that
$\Phi=\left[ \phi_0, \cdots, \phi_K \right] \in \Re^{\left| \state \right| \left| \action \right| \times K}$,
and $\Qspace=\Qspace_0^H$ with
$\Qspace_0 = \left \{ \Phi \theta: \, \theta \in \Re^K, \| \theta \|_0 \le K_0 \right \}$,
where $\| \theta \|_0$ is the $L_0$-``norm" of $\theta$ and $K_0 \ll K$.
\end{itemize}

It is worth mentioning that OCP, as we have defined it, assumes that an action $a$ maximizing $\sup_{Q \in \Qspace_{\mathcal{C}}} Q_t(x_{j,t},a)$
exists in each iteration.  
Note that this assumption always holds if the action space $\action$ is finite, and
it is not difficult to modify the algorithm so that it addresses cases where this is not true.  But we have
not presented the more general form of OCP in order to avoid complicating this paper.

Finally, we compare OCP with some classical reinforcement learning algorithms.
It is worth mentioning that in the finite state/action {\it tabula rasa} case, 
OCP is equivalent to the Q-learning algorithm with \emph{learning rate} $1$ and initial Q-value 
$Q_t (x,a) =\infty$, $\forall (x,a,t)$.
Please refer to the appendix for the justification of this argument.
On the other hand, in the linear generalization/approximation case with $\Qspace = \text{span}(\phi_1,\ldots,\phi_K)^H$, 
OCP is very different from the classical approaches where the weights are estimated 
using either temporal-difference learning (e.g. Q-learning with linear approximation) or least squares (e.g. least-squares value iteration).

\section{Sample Efficiency of Optimistic Constraint Propagation} \label{sec:sample_efficiency}

We now establish results concerning the sample efficiency (performance) of OCP. 
Our results bound the $\epsilon$-suboptimal 
sample complexities of OCP for appropriate choices of $\epsilon$.
Obviously, these sample complexity bounds
must depend on the complexity of the hypothesis class.  As such, we begin by defining the
eluder dimension, as introduced in \citet{Russo2013}, which is the notion of hypothesis class complexity we will use.

\subsection{Eluder Dimension} \label{sec:eluder_dimension}

Let $\SAT = \{(x,a,t) : x \in \state, a \in \action, t =0,\ldots,H-1\}$ 
be the set of all state-action-period triples, and let $\Qspace$ to denote a nonempty set of functions 
mapping $\SAT$ to $\Re$.  For all $(x,a,t) \in \SAT$ and $\subSAT \subseteq \SAT$, $(x,a,t)$ is said
to be {\it dependent} on $\subSAT$ with respect to $\Qspace$ if any pair of functions $Q,\tilde{Q} \in \Qspace$ that are equal
on $\subSAT$ are equal at $(x,a,t)$.  Further, $(x,a,t)$ is said to be {\it independent} of $\subSAT$ 
with respect to $\Qspace$ if $(x,a,t)$ is not dependent on $\subSAT$ with respect to $\Qspace$.

The {\it eluder dimension} $\dimM{\Qspace}$ of $\Qspace$
is the length of the longest sequence of elements in $\SAT$ such that every element is 
independent of its predecessors.
Note that $\dimM{\Qspace}$ can be zero or infinity, and it is straightforward to show that 
if $\Qspace_1 \subseteq \Qspace_2$ then $\dimM{\Qspace_1} \le \dimM{\Qspace_2}$.
Based on results of \citet{Russo2013}, we can characterize the eluder dimensions of various 
hypothesis classes presented in the previous section.
\begin{itemize}
\item \textbf{Finite state/action tabula rasa case.} If $\Qspace = \Re^{|\state| \cdot |\action| \cdot H}$, then
$\dimM{\Qspace} = |\state| \cdot |\action| \cdot H$.
\item \textbf{Polytopic prior constraints.}  If $\Qspace$ is a polytope of dimension $d$ in $\Re^{|\state| \cdot |\action| \cdot H}$, 
then $\dimM{\Qspace} = d$.
\item \textbf{Linear systems with quadratic cost (LQ).} If $\Qspace_0$ is the set of positive semidefinite quadratics with domain
$\Re^{m + n}$ and $\Qspace = \Qspace_0^H$, then $\dimM{\Qspace} = (m + n + 1)(m+n) H / 2$.
\item \textbf{Finite hypothesis space.} If $|\Qspace|< \infty$, then $\dimM{\Qspace} \leq |\Qspace|-1$.
\item \textbf{Linear combination of features.} If $\Qspace = \Qspace_0^H$ with $\Qspace_0 = \text{span}(\phi_1,\ldots,\phi_K)$,
then $\dimM{\Qspace} \leq K H$.
\item \textbf{State aggregation.} This is a special case of a linear combination of features.  If $\Qspace = \Qspace_0^H$,
and $\Qspace_0$ is the span of indicator functions for $K$ partitions of the state-action space, then $\dimM{\Qspace} \leq K H$.
\item \textbf{Sigmoid.} If $\Qspace = \Qspace_0^H$ with $\Qspace_0 = \left\{\psi\left(\sum_k \theta_k \phi_k(\cdot)\right): \theta \in \Re^K\right\}$,
then $\dimM{\Qspace} \leq K H$.
\item \textbf{Sparse linear combination of features.}
If $\Qspace=\Qspace_0^H$ with $\Qspace_0 = \left \{ \Phi \theta: \, \theta \in \Re^K, \| \theta \|_0 \le K_0 \right \}$ and 
$2K_0 \le \min \{|\state||\action|, K \}$, and
any $2K_0 \times 2K_0$ submatrix of $\Phi$ has full rank, then
$\dimM{\Qspace} \leq 2 K_0 H$. We will establish this eluder dimension bound in the appendix. 
\end{itemize}

\subsection{Learning with a Coherent Hypothesis Class} \label{sec:coherent_hypothesis_class}

We now present results that apply when OCP is presented with a coherent hypothesis class; that is, where $Q^* \in \Qspace$.
We refer to such cases as coherent learning cases.
Our first result establishes that OCP can deliver less than optimal performance in no more than $\dimM{\Qspace}$ episodes.
\begin{theorem}
For any system $\mdp=\left(\state, \action, H, F, R, S\right)$, if OCP is applied with $Q^* \in \Qspace$, then 
$|\{j : R^{(j)} < V^*_0(x_{j,0})\}| \leq \dimM{\Qspace}$.
\label{the:sample_efficiency}
\end{theorem}
That is, Theorem \ref{the:sample_efficiency} bounds the $0$-suboptimal sample complexity of OCP
in coherent learning cases.
This theorem follows from an ``exploration-exploitation lemma'' (Lemma \ref{lemma:ee1}), which 
asserts that in each episode, OCP either delivers optimal reward (exploits) 
or introduces a constraint that reduces the eluder dimension of the hypothesis class by one (explores). 
Consequently, OCP will experience sub-optimal performance in at most $\dimM{\Qspace}$ episodes.
We outline the proof of Theorem \ref{the:sample_efficiency} at the end of this subsection
and the detailed analysis is provided in the appendix.
An immediate corollary bounds regret.
\begin{corollary}
For any $\overline{R}$, any system $\mdp=\left(\state, \action, H, F, R, S\right)$ with $\sup_{(x,a,t)} |R_t(x,a)| \leq \overline{R}$, and any $T$, 
if OCP is applied with $Q^* \in \Qspace$, then $\text{\rm Regret}(T) \leq 2 \overline{R} H \dimM{\Qspace}$.
\label{the:regret}
\end{corollary}

Note the regret bound in Corollary \ref{the:regret} does not depend on time $T$, thus, it is an
$\BO{1}$ bound. 
Furthermore, this regret bound is linear in $\Rmax$, $H$ and $\dimM{\Qspace}$.
Thus, if $\dimM{\Qspace}$ does not depend on $|\state|$ or $|\action|$, then this regret bound also does not depend on
$|\state|$ or $|\action|$.
The following result demonstrates that the bounds of the above theorem and corollary are sharp.
\begin{theorem}
For any $\Rmax \geq 0$, any $K, H'=1,2,\cdots$ and any reinforcement learning algorithm $\RL$ that takes as input a
state space, an action space, a horizon and a coherent hypothesis class,
there exist a system $\mdp=(\state, \action, H, F, R, S)$ and a hypothesis class $\Qspace$ satisfying
(1) $\sup_{(x,a,t)} |R_t(x,a)| \leq \Rmax$, (2) $H=H'$, (3) $\dimM{\Qspace}=K$ and (4) $Q^* \in \Qspace$
such that if we apply $\RL$ to $\mdp$ with input $(\state,\action, H, \Qspace)$, then
$|\{j : R^{(j)} < V^*_0(x_{j,0})\}| \geq \dimM{\Qspace}$ and $\sup_T \text{\rm Regret}(T) \geq 2 \overline{R} H \dimM{\Qspace}$.
\label{the:tight}
\end{theorem}

A constructive proof of these lower bounds is provided at the end of this subsection.
Following our discussion in previous sections, we discuss several interesting contexts in which
the agent knows a coherent hypothesis class $\Qspace$ with finite eluder dimension.
\begin{itemize}
\item \textbf{Finite state/action tabula rasa case.} If we apply OCP in this case, then it will deliver sub-optimal performance in at most
$|\state| \cdot |\action| \cdot H$ episodes. Furthermore, if $\sup_{(x,a,t)} |R_t(x,a)| \leq \overline{R}$, then for any $T$, 
$\text{\rm Regret}(T) \leq 2 \overline{R} |\state|  |\action|  H^2$.
\item \textbf{Polytopic prior constraints.} If we apply OCP in this case, then it will deliver sub-optimal performance in at most
$d$ episodes. Furthermore, if $\sup_{(x,a,t)} |R_t(x,a)| \leq \overline{R}$, then for any $T$, 
$\text{\rm Regret}(T) \leq 2 \overline{R}  H d$.
\item \textbf{Linear systems with quadratic cost (LQ).} If we apply OCP in this case, then it will deliver sub-optimal performance in at most
$(m + n + 1)(m+n) H / 2$ episodes.
\item \textbf{Finite hypothesis class case.} Assume that the agent has prior knowledge that $Q^* \in \Qspace$, where $\Qspace$ is a finite
hypothesis class. If we apply OCP in this case, then it will deliver sub-optimal performance in at most $|\Qspace|-1$ episodes.
Furthermore, if $\sup_{(x,a,t)} |R_t(x,a)| \leq \overline{R}$, then for any $T$, 
$\text{\rm Regret}(T) \leq 2 \overline{R}  H  \left[ |\Qspace|-1\right]$.
\item \textbf{Linear combination of features.} Assume that $Q^* \in \Qspace = \Qspace_0^H$ with $\Qspace_0 = \text{span}(\phi_1,\ldots,\phi_K)$.
If we apply OCP in this case, then it will deliver sub-optimal performance in at most 
$KH$ episodes. Furthermore, if $\sup_{(x,a,t)} |R_t(x,a)| \leq \overline{R}$,
then for any $T$, 
$\text{\rm Regret}(T) \leq 2 \overline{R} K  H^2$.
Notice that this result can also be derived based on the KWIK online regression with deterministic linear functions (see \citep{li2011knows}).
\item \textbf{Sparse linear combination case.}
Assume that the agent has prior knowledge that
$Q^* \in \Qspace$, where 
$\Qspace=\left \{ \Phi \theta: \, \theta \in \Re^K, \| \theta \|_0 \le K_0 \right \}^H$ and $2K_0 \le \min \{|\state||\action|, K \}$, and
any $2K_0 \times 2K_0$ submatrix of $\Phi$ has full rank.
If we apply OCP in this case, then it will deliver sub-optimal performance in at most
$2 K_0 H$ episodes.
Furthermore, if $\sup_{(x,a,t)} |R_t(x,a)| \leq \overline{R}$, then for any $T$, 
$\text{\rm Regret}(T) \leq 4 \overline{R}  K_0  H^2$.
\end{itemize}

Before proceeding, it is worth pointing out that one key feature of OCP, which distinguishes it from other reinforcement learning algorithms and makes it sample efficient when presented with a coherent hypothesis class, is that it updates the feasible set of candidates for $Q^*$  in a conservative manner that never rues out $Q^*$ and always uses optimistic estimates from this feasible set to guide action.

\subsubsection{Sketch of Proof for Theorem \ref{the:sample_efficiency}}
We start by defining some useful notations.
Specifically, we use $\CC_j$ to denote the $\CC$ in episode $j$
 to distinguish $\CC$'s in different episodes, and use $\sat$ as a shorthand notation for
a state-action-time triple $(x,a,t)$.
We first prove that if  $Q^* \in \Qspace$, then each constraint appended to
$\CC$ does not rule out $Q^*$, and thus we have $Q^* \in \Qspace_{\CC_{j}}$ for 
any $j=0,1,\cdots$.
\begin{lemma}
\label{lemma:tech0}
If $Q^* \in \Qspace$, then (a) $Q^* \in \Qspace_{\CC_{j}}$ for 
all $j=0,1,\cdots$, and (b) 
$L_{j,t} \leq Q^*_t (x_{j,t}, a_{j,t}) \leq U_{j,t}$ for all $t$ and all 
$j=0,1,\cdots$.
\end{lemma}
Please refer to the appendix for the proof of Lemma \ref{lemma:tech0}.
Notice that Lemma \ref{lemma:tech0}(b) implies that no constraints are conflicting if
$Q^* \in \Qspace$ since
$Q^*$ satisfies all the constraints.
For any episode $j=0,1,\cdots$, we define $\SAT_j$ and $t_j^*$ by Algorithm \ref{alg:auxiliary}.

\begin{algorithm}[th]
\caption{Definition of $\SAT_j$ and $t_j^*$}
\label{alg:auxiliary}
\begin{algorithmic}
\State Initialize $\SAT_0 \leftarrow \varnothing$
\For{$j=0,1,\cdots$}
\State Set $t_j^* \leftarrow \NULL$
\If{$\exists t=0,1,\cdots, H-1$ s.t. $(x_{j,t}, a_{j,t}, t)$ is independent of $\SAT_j$ with respect to $\Qspace$}
\State Set
\BE
\hspace{0.7cm}
t_j^* & \leftarrow & \textrm{last period $t$ in episode $j$ s.t. $(x_{j,t}, a_{j,t}, t)$ is independent of $\SAT_j$ with respect to $\Qspace$} \nonumber
\EE
\hspace{1cm} and
$
\SAT_{j+1} \leftarrow \left[ \SAT_j ,  (x_{j,t_j^*}, a_{j,t_j^*}, t_j^*) \right]
$
\Else
\State Set $\SAT_{j+1} \leftarrow \SAT_j$
\EndIf
\EndFor
\end{algorithmic}
\end{algorithm}
Note that by definition, in each episode $j$, $\SAT_j$ is a sequence (ordered set) of elements in $\SAT$. Furthermore, each element in $\SAT_j$ is independent of its predecessors.
Moreover, if $t_j^* \neq \NULL$, then it is the last period in episode $j$ s.t. $(x_{j,t}, a_{j,t}, t)$ is independent of $\SAT_j$ with respect to $\Qspace$.
As we will show in the analysis, if $t_j^* \neq \NULL$, another interpretation of $t^*_j$ is that it is the first period (in backward order) in episode $j$ 
when the value of a new state-action-period triple is learned perfectly.
Based on the notions of $\SAT_j$ and $t_j^*$, we have the following technical lemma:

\begin{lemma}
$\forall j=0,1,\cdots$ and $\forall t=0,1,\cdots, H-1$, we have
\begin{enumerate}[label=\rm{(\alph*)}]
\item $\forall \sat \in \SAT_j$ and $\forall Q \in \Qspace_{\CC_{j}}$, we have $Q(\sat)=Q^*(\sat)$.
\item If $(x_{j,t}, a_{j,t}, t)$ is dependent on $\SAT_j$ with respect to $\Qspace$, then (1) $a_{j,t}$ is optimal and (2)
$Q_t (x_{j,t}, a_{j,t})=Q^*_t (x_{j,t}, a_{j,t})=V^*_t (x_{j,t})$,  $\forall Q \in \Qspace_{\CC_{j}}$.
\end{enumerate}
\label{lemma:tech1}
\end{lemma}

Please refer to the appendix for the proof of Lemma \ref{lemma:tech1}.
Based on Lemma \ref{lemma:tech1}, we have the following exploration/exploitation lemma, which states that in each episode
$j$, OCP algorithm either achieves the optimal reward (exploits), or updates
$\Qspace_{\CC_{j+1}}$ based on the Q-value at an independent state-action-time triple (explores).

\begin{lemma}
For any $j=0,1,\cdots$, 
if $t_j^* \neq \NULL$, then $(x_{j,t_j^*}, a_{j,t_j^*}, t_j^*)$ is independent of $\SAT_j$, 
$|\SAT_{j+1}|=|\SAT_j|+1$ and 
$Q_{t_j^*} (x_{j,t_j^*}, a_{j,t_j^*})=Q^*_{t_j^*} (x_{j,t_j^*}, a_{j,t_j^*})$
$\forall Q \in \Qspace_{\CC_{j+1}}$ \rm{(Exploration)}.
Otherwise, if $t_j^*=\NULL$, then $R^{(j)}=V^*_0 (x_{j,0})$ \rm{(Exploitation)}.
\label{lemma:ee1}
\end{lemma}

Theorem \ref{the:sample_efficiency} follows from Lemma \ref{lemma:ee1}. Please refer to the appendix for the detailed proofs for
Lemma \ref{lemma:ee1} and Theorem \ref{the:sample_efficiency}.

\subsubsection{Constructive Proof for Theorem \ref{the:tight}}
\label{sec:constructive_proof_thm2}
We start by defining some useful terminologies and notations.
First, for any state space $\state$, any time horizon $H=1,2,\cdots$, any action space $\action$,
and any hypothesis class $\Qspace$,
we use $\mdpset \left( \state, \action, H, \Qspace \right)$ to denote the set of all finite-horizon deterministic system $\mdp$'s with
state space $\state$, action space $\action$, horizon $H$ and $Q^* \in \Qspace$.
Notice that for any reinforcement learning algorithm that takes $\state$, $\action$, $H$, $\Qspace$ as input, and knows that
$\Qspace$ is a coherent hypothesis class,
$\mdpset \left( \state, \action, H, \Qspace \right)$ is the set of all finite-horizon deterministic systems that are consistent with the algorithm's prior information.

We provide a constructive proof for Theorem \ref{the:tight} by considering a scenario in which an adversary
adaptively chooses a deterministic system $\mdp \in \mdpset \left( \state, \action, H, \Qspace \right)$.
Specifically, we assume that
\begin{itemize}
\item At the beginning of each episode $j$, the adversary adaptively chooses the initial state $x_{j,0}$.
\item At period $t$ in episode $j$, the agent first chooses an action $a_{j,t} \in \action$ based on some RL algorithm\footnote{In general, the RL algorithm
can choose actions randomly. If so, all the results in Section \ref{sec:constructive_proof_thm2} hold on the realized sample path.}, and then the
adversary adaptively chooses a set of state-action-time triples $\SAT_{j,t} \subseteq \SAT$ and specifies the rewards and
state transitions on $\SAT_{j,t}$, subject to the constraints that (1) $(x_{j,t}, a_{j,t}, t) \in \SAT_{j,t}$ and (2)
these adaptively specified rewards and
state transitions must be consistent with the agent's prior knowledge
and past observations.
\end{itemize}
We assume that the adversary's objective is to maximize the number of episodes in which the agent achieves sub-optimal rewards.
Then we have the following lemma:
\begin{lemma}
$\forall H,K=1,2,\cdots$ and $\forall \Rmax \ge 0$, there exist a state space $\state$, an action space $\action$ and
a hypothesis class $\Qspace $ with $\dimM{\Qspace}=K$ such that no matter how the agent adaptively chooses actions,
the adversary can adaptively choose an $\mdp \in \mdpset \left( \state, \action, H, \Qspace \right)$
with $\sup_{(x,a,t)} |R_t(x,a)| \le \Rmax$ such that the agent will achieve sub-optimal rewards in at least $K$ episodes, and $\sup_{T} \reg(T) \geq 2 \Rmax H K$.
\label{lemma:tight}
\end{lemma}
Since the fact that an adversary can adaptively choose a ``bad" deterministic system simply implies that such a system exists,
thus, 
Theorem \ref{the:tight} follows directly from Lemma \ref{lemma:tight}.\\
\proof{Proof for Lemma \ref{lemma:tight}}
We provide a constructive proof for Lemma \ref{lemma:tight}. Specifically,
$\forall H,K=1,2,\cdots$ and $\forall \Rmax \ge 0$, we construct the state space
as $\state=\left \lbrace 1,2, \cdots, 2K \right \rbrace$, and the action space 
as $\action = \{ 1,2 \}$.
Recall that $\SAT=\left \lbrace (x,a,t) : \, x\in \state, t=0,1,\cdots, H-1, \textrm{ and } a\in \action \right \rbrace$, 
thus, for $\state$ and $\action$ constructed above, we have
$|\SAT|=4KH$. Hence, $Q^*$, the optimal Q-function, can be represented as a vector in $\Re^{4KH}$.

Before specifying the hypothesis class $\Qspace$, we first define a matrix  $\Phi \in \Re^{4KH \times K}$ as follows.
$\forall (x,a,t) \in \SAT$, let $\Phi (x,a,t) \in \Re^K$ denote the row of $\Phi$ corresponding to the state-action-time triple
$(x,a,t)$, we construct $\Phi (x,a,t)$ as:
\BE
\Phi (x,a,t)= \left \lbrace
\begin{array}{ll}
(H-t) \mathbf{e}_k & \textrm{if $x=2k-1$ for some $k=1,\cdots,K$, $a=1,2$ and $t=1,\cdots,H-1$} \\
-(H-t) \mathbf{e}_k & \textrm{if $x=2k$ for some $k=1,\cdots,K$, $a=1,2$ and $t=1,\cdots,H-1$} \\
H \mathbf{e}_k & \textrm{if $x=2k-1$ or $2k$ for some $k=1,\cdots,K$, $a=1$ and $t=0$} \\
-H \mathbf{e}_k & \textrm{if $x=2k-1$ or $2k$ for some $k=1,\cdots,K$, $a=2$ and $t=0$} 
\end{array}
\right.  \label{theorem3:phi}
\EE
where $\mathbf{e}_k \in \Re^K$ is a (row) indicator vector with a one at index $k$ and zeros
everywhere else. 
Obviously, $\mathrm{rank} (\Phi)=K$. We choose $\Qspace=\spann \left[ \Phi \right ]$, thus 
$\dimM{\Qspace}=\dim \left(  \spann \left[ \Phi \right ] \right)=\mathrm{rank} (\Phi)=K$.

Now we describe how the adversary adaptively chooses a finite-horizon deterministic system $\mdp \in \mdpset \left( \state, \action, H, \Qspace \right)$:
\begin{itemize}
\item For any $j=0,1,\cdots$, at the beginning of episode $j$, the adversary chooses the initial state in that episode as
$x_{j,0}= (j \bmod K) \times 2 +1$. 
That is, $x_{0,0}=x_{K,0}=x_{2K,0}=\cdots=1$, $x_{1,0}=x_{K+1,0}=x_{2K+1,0}=\cdots=3$, etc.
\item Before interacting with the agent, the adversary chooses the following system function $F$\footnote{More precisely, in this constructive proof, the adversary does not need to adaptively choose
the system function $F$. He can choose $F$ beforehand.}:
\[
F_t (x,a)= 
\left \lbrace
\begin{array}{ll}
2k-1 & \textrm{if $t=0$, $x=2k-1$ or $2k$ for some $k=1,\cdots, K$, and $a=1$} \\
2k & \textrm{if $t=0$, $x=2k-1$ or $2k$ for some $k=1,\cdots, K$, and $a=2$} \\
x & \textrm{if $t=1,\cdots, H-2$ and $a=1,2$}
\end{array}
\right. .
\]
The state transition is illustrated in Figure \ref{f2}.
\begin{figure}
\centering
\includegraphics[scale=0.6]{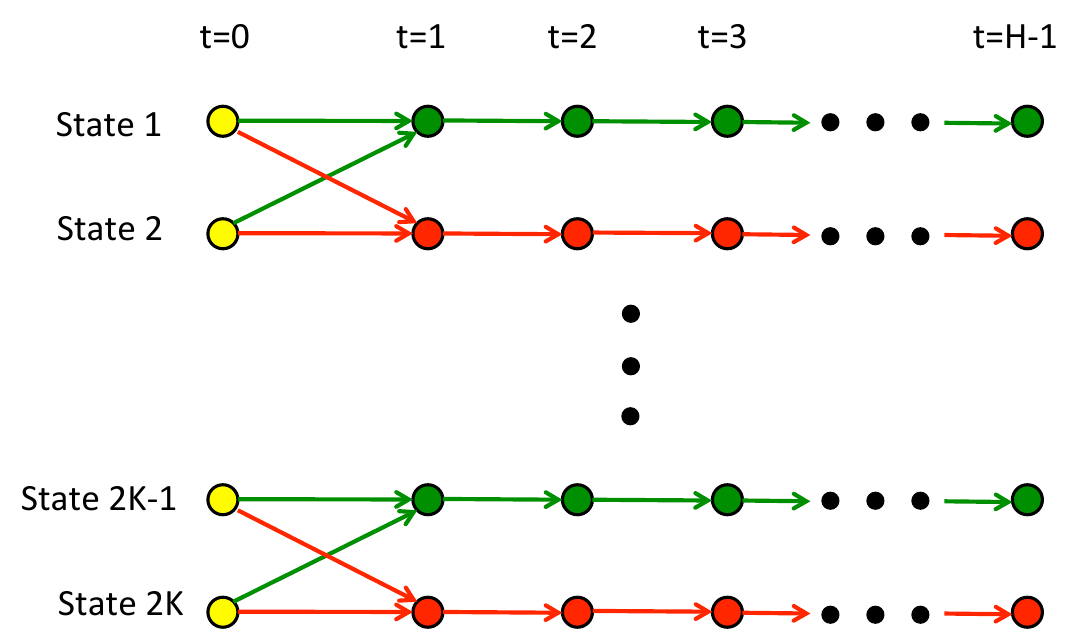}
\caption{Illustration of the state transition}
\label{f2}
\end{figure}

\item In episode $j=0,1,\cdots, K-1$, the adversary adaptively chooses the reward function $R$ as follows.
If the agent takes action $1$ in period $0$ in episode $j$ at initial state $x_{j,0}=2j+1$, then the adversary set
$R_0 (2j+1, 1) =R_0 (2j+2, 1) = R_t (2j+1,1) = R_t (2j+1,2)=-\Rmax$ and
$R_0 (2j+1, 2) =R_0 (2j+2, 2) = R_t (2j+2,1) = R_t (2j+2,2)=\Rmax$,
$\forall t=1,2,\cdots, H-1$. Otherwise (i.e. if the agent takes action $2$ in period $0$ in episode $j$), then the adversary set
$R_0 (2j+1, 1) =R_0 (2j+2, 1) = R_t (2j+1,1) = R_t (2j+1,2) = \Rmax $
and
$R_0 (2j+1, 2) =R_0 (2j+2, 2) = R_t (2j+2,1) = R_t (2j+2,2) = -\Rmax$.
Notice that the adversary completes the construction of the deterministic system $\mdp$ at the end of episode $K-1$.
\end{itemize}

Note that for the constructed deterministic system $\mdp$, we have $Q^* \in \Qspace$.
Specifically, it is straight forward to see that 
$Q^*=\Phi \theta^*$, where $\theta^* \in \Re^K$, and
$\theta^*_k$, the $k$th element of $\theta$, is defined as
$\theta^*_k = -\Rmax$ if $a_{k-1,0}=1$ and
$\theta^*_k = \Rmax$ if $a_{k-1,0}=2$,
for any $k=1,2,\cdots, K$. Thus, the constructed deterministic system $\mdp \in \mdpset \left( \state, \action, H, \Qspace \right)$.

Finally, we show that the constructed deterministic system $\mdp$ satisfies Lemma \ref{lemma:tight}.
Obviously, we have $|R_t(x,a)| \le \Rmax$, $\forall (x,a,t) \in \SAT$. Furthermore, note that the agent achieves sub-optimal rewards in the first
$K$ episodes, thus, he will achieve sub-optimal rewards in at least $K$ episodes. In addition, the cumulative regret in the first $K$ episodes is
$2 K H \Rmax$, thus,  $\sup_T \reg(T) \ge 2K H \Rmax$.
\endproof

\subsection{Agnostic Learning in State Aggregation Case} \label{sec:agnostic_learning}

As we have discussed in Section \ref{sec:algorithm}, OCP can also be applied in agnostic learning cases,
where $Q^*$ may not lie in $\Qspace$. For such cases, the performance of
OCP should depend on not only the complexity of $\Qspace$, but also the distance between
$\Qspace$ and $Q^*$.
In this subsection, 
we present results when OCP is applied in a special agnostic learning case,
where $\Qspace$ is the span of pre-specified indicator functions over disjoint
subsets. We henceforth refer to this case as the state aggregation case.

Specifically, we assume that for any $t=0,1,\cdots,H-1$,
the state-action space at period $t$,
$\SAT_t= \left \lbrace (x,a, t): \, x \in \state, a \in \action \right \rbrace$,
can be partitioned into $K_t$ disjoint subsets $\SAT_{t,1}, \SAT_{t,2}, \cdots, \SAT_{t, K_t}$,
and use $\phi_{t,k}$ to denote the indicator function for partition
$\SAT_{t,k}$ (i.e. $\phi_{t,k} (x,a,t)=1$ if $(x,a,t) \in \SAT_{t,k}$, and $\phi_{t,k}(x,a,t)=0$ otherwise). 
We define $K=\sum_{t=0}^{H-1} K_t$, and
$\Qspace$ as
\BE
\Qspace= \spann \left \lbrace  \phi_{0,1}, \phi_{0,2}, \cdots, \phi_{0, K_0}, \phi_{1,1}, \cdots, \phi_{H-1, K_{H-1}} \right \rbrace . \label{eqn:Qspace}
\EE
Note that $\dimM{\Qspace}=K$.
We define the distance between $Q^*$ and the hypothesis class $\Qspace$ as
\BE
\rho=\min_{Q \in \Qspace} \| Q-Q^* \|_{\infty}=\min_{Q \in \Qspace} \sup_{(x,a,t)} |Q_t (x,a)-Q^*_t(x,a)|. \label{eqn:rho}
\EE

The following result establishes that with $\Qspace$ and $\rho$ defined above, the performance loss of OCP
is larger than $2 \rho H (H+1)$ in at most $K$ episodes.

\begin{theorem}
For any system $\mdp=\left(\state, \action, H, F, R, S\right)$, if OCP is applied with $\Qspace$ defined in Eqn(\ref{eqn:Qspace}), 
then 
$ |\{ j: R^{(j)} < V^*_0 (x_{j,0})- 2 \rho H (H+1)\}| \le K $,
where $K$ is the number of partitions and 
$\rho$ is defined in Eqn(\ref{eqn:rho}).
\label{the:state_aggregation}
\end{theorem}
That is, Theorem \ref{the:state_aggregation} bounds the $2 \rho H (H+1)$-suboptimal sample complexity of OCP
in the state aggregation case.
Similar to Theorem \ref{the:sample_efficiency},
this theorem also follows from an ``exploration-exploitation lemma" (Lemma \ref{state_aggregation:lemma3}), which asserts that in each episode,
OCP either delivers near-optimal reward (exploits), or approximately determines $Q^*_t (x,a)$'s for 
all the $(x,a,t)$'s in a disjoint subset (explores).
We outline the proof for Theorem \ref{the:state_aggregation} at the end of this subsection,
and the detailed analysis is provided in the appendix.
An immediate corollary bounds regret.

\begin{corollary}
For any $\Rmax \geq 0$, any system $\mdp=\left(\state, \action, H, F, R, S\right)$ with $\sup_{(x,a,t)} |R_t(x,a)| \leq \Rmax$, and any time $T$, 
if OCP is applied with $\Qspace$ defined in Eqn(\ref{eqn:Qspace}), then
$\text{\rm Regret}(T) \leq 2 \Rmax K H + 2 \rho (H+1) T$, where $K$ is the number of partitions and 
$\rho$ is defined in Eqn(\ref{eqn:rho}).
\label{the:state_aggregation_regret}
\end{corollary}
Note that the regret bound in Corollary \ref{the:state_aggregation_regret} is $\BO{T}$,
and the coefficient of the linear term is $2 \rho (H+1)$. Consequently, if $Q^*$ is close to $\Qspace$, then the 
regret will increase slowly with $T$.
Furthermore, the regret bound in Corollary \ref{the:state_aggregation_regret}
does not directly depend on $|\state|$ or $|\action|$.

We further notice that the threshold performance loss in Theorem \ref{the:state_aggregation} is
$\BO{\rho H^2}$. The following proposition provides a condition under which the 
performance loss in one episode is $\BO{\rho H}$.

\begin{proposition}
For any episode $j$, if 
$ \Qspace_{\mathcal{C}} \subseteq \left \lbrace Q \in \Qspace:\, L_{j,t} \le Q_t (x_{j,t}, a_{j,t}) \le U_{j,t} \right \rbrace $, $\forall t=0,\cdots, H-1$, 
then we have
$ V_0^* \left( x_{j,0} \right) - R^{(j)} \le 6 \rho H = \BO{\rho H}$. 
\label{the:performance_loss}
\end{proposition}

That is, if all the new constraints in an episode are redundant,
then the performance loss in that episode is $\BO{\rho H}$.
Note that if the condition for Proposition \ref{the:performance_loss}
holds in an episode, then $\Qspace_{\mathcal{C}}$ will not be modified
at the end of that episode.
Furthermore, if the system has a fixed initial state and the condition for Proposition \ref{the:performance_loss}
holds in one episode, then it will hold in all the subsequent episodes, and consequently,
the performance losses in all the subsequent episodes are $\BO{\rho H}$.

It is worth mentioning that the sample complexity bound and the regret bounds in this subsection are derived under the assumption that the partitions of the state-action spaces are given. An important problem in practice is how to choose the optimal number $K$ of the state-action partitions. There are many approaches to choose $K$, and one approach is to formulate it as
a regret bound optimization problem. Specifically, assume that for any $K \geq H$, $\Qspace(K)$ is the hypothesis class the agent constructs with $K$ partitions. Let $\rho(K)$ be a known upper bound on the distance
$\min_{Q \in \Qspace(K)} \| Q-Q^*\|_{\infty}$. Then from Corollary~\ref{the:state_aggregation_regret} , 
$\text{\rm Regret}(T) \leq 2 \Rmax K H + 2 \rho(K) (H+1) T$.
Hence, the problem of choosing an optimal $K$ can be formulated as
\[
\min_{K \geq H}\, 2 \bar{R} K H + 2 \rho(K) (H+1) T,
\]
which can be efficiently solved by line search. Notice that whether or not the optimal $K$ depends on $|\mathcal{S}| |\mathcal{A}| $, and/or how it grows with $|\mathcal{S}| |\mathcal{A} |$, 
depends on if and how $\rho(K)$ depends on $|\mathcal{S}| |\mathcal{A}|$. That is, it depends on the agent's capability to construct a good hypothesis class $\mathcal{Q}(K)$ for a given $K$, which in turn might depend on the agent's prior knowledge about the problem.

\subsubsection{Sketch of Proof for Theorem \ref{the:state_aggregation} and Proposition \ref{the:performance_loss}}

We start by briefly describing how constraint selection algorithm updates
$\Qspace_{\CC}$'s for the function class $\Qspace$ specified in Eqn(\ref{eqn:Qspace}).
Specifically, let $\theta_{t,k}$ denote the coefficient of the indicator function
$\phi_{t,k}$, $\forall (t,k)$.
Assume that $(x,a,t)$ belongs to partition
$\SAT_{t,k}$, then, with $\Qspace$ specified in Eqn(4.1),
$L \leq Q_t(x,a) \leq U$ is a constraint on and only on
$\theta_{t,k}$, and is equivalent to
$L \leq \theta_{t,k} \leq U$.
By induction, it is straightforward to see in episode $j$,
$\Qspace_{\CC_j}$ can be represented as
\BE
\Qspace_{\CC_j}=
\left \{  
\theta \in \Re^K : \,  \underline{\theta}^{(j)}_{t,k} \leq \theta_{t,k} \leq \overline{\theta}^{(j)}_{t,k}, \, \forall (t,k)
\right \},
\EE
for some  $\underline{\theta}^{(j)}_{t,k}$'s and $\overline{\theta}^{(j)}_{t,k}$'s.
Note that  $\underline{\theta}^{(j)}_{t,k}$ can be $-\infty$ and $\overline{\theta}^{(j)}_{t,k}$
can be $\infty$, and when $j=0$, $\overline{\theta}^{(0)}_{t,k}=\infty$
and $\underline{\theta}^{(0)}_{t,k}=-\infty$.
Furthermore, from the constraint selection algorithm,
$\overline{\theta}^{(j)}_{t,k}$ is monotonically non-increasing in $j$,  $\forall (t,k)$.
Specifically, if OCP adds a new constraint $ L \leq \theta_{t,k} \leq U$ on $\theta_{t,k}$ in episode $j$, we have
$\overline{\theta}^{(j+1)}_{t,k}=\min \{ \overline{\theta}^{(j)}_{t,k}, U  \}$;
otherwise, $\overline{\theta}^{(j+1)}_{t,k}=\overline{\theta}^{(j)}_{t,k}$.
Thus, if $\overline{\theta}^{(j)}_{t,k} < \infty$, then $\overline{\theta}^{(j')}_{t,k} < \infty$, $\forall j' \geq j$.

For any episode $j$, we define $\Qopt_{j}$, the optimistic Q-function in episode $j$, as
\BE
\Qopt_{j,t}(x,a) = \sup_{Q \in \Qspace_{\CC_j}} Q_t (x,a), \quad \forall (x,a,t)\in \SAT.
\EE
Similarly, $\Qpes_{j}$, the pessimistic 
Q-function in episode $j$, is defined as
\BE
\Qpes_{j,t}(x,a) = \inf_{Q \in \Qspace_{\CC_j}} Q_t (x,a), \quad \forall (x,a,t)\in \SAT.
\EE
%
%
%
Clearly, if $(x,a,t) \in \SAT_{t,k}$, then we have
$\Qopt_{j,t}(x,a) =\overline{\theta}^{(j)}_{t,k}$,
and 
$\Qpes_{j,t}(x,a) =\underline{\theta}^{(j)}_{t,k}$. Moreover, $(x,a,t)$'s
in the same partition have the same optimistic and pessimistic Q-values.

It is also worth pointing out that by definition of $\rho$, if $(x,a,t)$ and
$(x',a',t)$ are in the same partition, then we have
$|Q^*_t (x,a)- Q^*_t (x',a')| \le 2 \rho$. To see it, let 
$\tilde{Q} \in \argmin_{Q \in \Qspace} \|Q-Q^* \|_{\infty}$, then we have
$| \tilde{Q}_t(x,a)-Q^*_t(x,a) | \le \rho$ and
$| \tilde{Q}_t(x',a')-Q^*_t(x',a') | \le \rho$.
Since $\tilde{Q} \in \Qspace$ and $(x,a,t)$ and
$(x',a',t)$ are in the same partition, we have
$\tilde{Q}_t(x,a)=\tilde{Q}_t(x',a')$.
Then from the triangular inequality, we have 
$|Q^*_t (x,a)- Q^*_t (x',a')| \le 2 \rho$.

The following lemma states that if $\Qopt_{j,t}(x,a)< \infty$, then it is ``close" to
$Q^*_t(x,a)$.
\begin{lemma}
$\forall (x,a,t)$ and $\forall j=0,1,\cdots$, if $\Qopt_{j,t}(x,a)< \infty$, then
$|\Qopt_{j,t} (x,a)-Q^*_t(x,a)| \leq 2 \rho (H-t)$.
\label{lemma:aggregation1}
\end{lemma}
Please refer to the appendix for the detailed proof of Lemma \ref{lemma:aggregation1}.
Based on this lemma, we have the following result:
\begin{lemma}
$\forall j=0,1,\cdots$, if
$\Qopt_{j,t} (x_{j,t}, a_{j,t}) < \infty$ for any $t=0,1,\cdots, H-1$, then we have
$
V^*_{0}(x_{j,0})- R^{(j)} \le 2 \rho H(H+1) = \BO{\rho H^2}
$.
Furthermore, if the conditions of Proposition \ref{the:performance_loss} hold,
then we have
$V^*_{0}(x_{j,0})- R^{(j)} \le 6 \rho H =O (\rho H)$. 
\label{lemma:aggregation2}
\end{lemma}
Please refer to the appendix for the detailed proof of Lemma \ref{lemma:aggregation2}.
Obviously, Proposition \ref{the:performance_loss} directly follows from Lemma \ref{lemma:aggregation2}.

For any $j=0,1,\cdots$, we define $t_j^*$ as the last period $t$
in episode $j$ s.t. 
$\Qopt_{j,t} (x_{j,t}, a_{j,t}) =\infty$.
If $\Qopt_{j,t} (x_{j,t}, a_{j,t}) < \infty$ for all $t=0,1,\cdots, H-1$, we define
$t_j^*=\NULL$. We then have the following lemma:
\begin{lemma}
$\forall j=0,1,\cdots$, 
if $t_j^* \neq \NULL$, then $\forall j' \leq j$, $\Qopt_{j',t_j^*} (x_{j,t_j^*}, a_{j,t_j^*}) = \infty$, and $\forall j' > j$,
$\Qopt_{j',t_j^*} (x_{j,t_j^*}, a_{j,t_j^*}) < \infty$ {\rm (Exploration)}.
Otherwise, if $t_j^* = \NULL$, then $V^*_{0}(x_{j,0})- R^{(j)} \le 2 \rho H(H+1)$ {\rm (Exploitation)}.
Furthermore,
$\sum_{j=0}^{\infty} \indicator [t_j^* \neq \NULL] \le K$, where $K$ is the number of partitions.
\label{state_aggregation:lemma3}
\end{lemma}
Again, please refer to the appendix for the proof of Lemma \ref{state_aggregation:lemma3}.
Note that Theorem \ref{the:state_aggregation} directly follows from Lemma \ref{state_aggregation:lemma3}.

\section{Computational Efficiency of Optimistic Constraint Propagation} \label{sec:computational_efficiency}

We now briefly discuss the computational complexity of OCP. 
As typical in the complexity analysis of optimization algorithms, we assume that basic operations include the
arithmetic operations, comparisons, and assignment, and measure computational complexity in terms of 
the number of basic operations (henceforth referred to as operations) per period.

 First, it is worth pointing out that for a general hypothesis class $\Qspace$ and general action space
$\action$, the per period computations of OCP can be intractable. This is because: 
\begin{itemize}
\item Computing $\sup_{Q \in \Qspace_{\mathcal{C}}} Q_t(x_{j,t}, a)$, $U_{j,t}$ and
$L_{j,t}$ requires solving a possibly intractable optimization problems.
\item Selecting an action that maximizes $\sup_{Q \in \Qspace_{\mathcal{C}}} Q_t(x_{j,t}, a)$ can be intractable.
\end{itemize}
Further, the number of constraints in $\mathcal{C}$, and with it the number of operations per period, can grow over time.

However, if $|\action|$ is tractably small and $\Qspace$ has some special structures (e.g. $\Qspace$ is a finite set or a linear subspace or, more generally a polytope),
then by discarding the ``redundant" constraints in $\mathcal{C}$, OCP with a variant of the constraint selection algorithm
%
will be computationally efficient, and the sample
efficiency results developed in Section \ref{sec:sample_efficiency} will still hold. Due to space limitations, we only discuss the scenario where 
$\Qspace$ is a polytope of dimension $d$. Note that the finite state/action {\it tabula rasa} case, the linear-quadratic case, and the 
state aggregation case
are all special cases of this scenario. 
Moreover, as we have discussed before, for the finite state/action {\it tabula rasa} case and the linear-quadratic case,
$Q^* \in \Qspace$.

Specifically, if $\Qspace$ is a polytope of dimension $d$ (i.e., within a $d$-dimensional subspace), then any $Q \in \Qspace$ can be represented by a weight vector $\theta \in \Re^d$, and
$\Qspace$ can be characterized by a set of linear inequalities of $\theta$. Furthermore, the new constraints of the form $L_{j,t} \leq Q_t (x_{j,t}, a_{j,t}) \leq U_{j,t}$
are also linear inequalities of $\theta$. Hence, in each episode, $\Qspace_{\mathcal{C}}$ is characterized by a polyhedron in $\Re^d$,
and
$\sup_{Q \in \Qspace_{\mathcal{C}}} Q_t(x_{j,t}, a)$, $U_{j,t}$ and
$L_{j,t}$ can be computed by solving linear programming (LP) problems. 
If we assume that each observed numerical value can be encoded by $B$ bits,
and LPs are solved by Karmarkar's algorithm \citep{Karmarkar1984}, then the following proposition bounds the 
computational complexity.
\begin{proposition}
If $\Qspace$ is a polytope of dimension $d$, each numerical value in the problem data or observed in the course of learning 
can be represented with $B$ bits, and OCP uses Karmarkar's algorithm
to solve linear programs, then the computational complexity of OCP is $\BO{ \left[ |\action|+|\mathcal{C}| \right] |\mathcal{C}| d^{4.5} B}$ operations per period.
\label{the:computational_complexity}
\end{proposition}
\proof{Proof}
Note that
OCP needs to perform the following computation in one period:
\begin{enumerate}
\item Construct $\Qspace_{\mathcal{C}}$ by constraint selection algorithm. This requires sorting $|\mathcal{C}|$ constraints by comparing their upper bounds and
positions in the sequence (with $\BO{|\mathcal{C}| \log |\mathcal{C}|}$ operations), and checking whether $\Qspace_{\mathcal{C}} \cap \mathcal{C}_{\tau} \neq \varnothing$ 
for $|\mathcal{C}|$ times. Note that checking whether $\Qspace_{\mathcal{C}} \cap \mathcal{C}_{\tau} \neq \varnothing$ requires solving an LP feasibility problem with 
$d$ variables and $\BO{|\mathcal{C}|}$ constraints.
\item Choose action $a_{j,t}$. Note that $\sup_{Q \in \Qspace_{\mathcal{C}}} Q_t(x_{j,t}, a)$ can be computed by solving an LP with $d$ variables and 
$\BO{|\mathcal{C}|}$ constraints, thus $a_{j,t}$ can be derived by solving $|\action|$ such LPs.
\item Compute the new constraint $L_{j,t} \leq Q_t (x_{j,t}, a_{j,t}) \leq U_{j,t}$. Note $U_{j,t}$ can be computed by solving
$|\action|$ LPs with $d$ variables and $\BO{|\mathcal{C}|}$ constraints, and $L_{j,t}$ can be computed by solving one LP with
$d$ variables and $\BO{|\mathcal{C}|+|\mathcal{A}|}$ constraints. 
\end{enumerate}
If we assume that each observed numerical value can be encoded by $B$ bits, and use Karmarkar's algorithm to solve LPs, then for an LP with
$d$ variables and $m$ constraints, the number of bits input to Karmarkar's algorithm is $\BO{mdB}$, and hence it requires $\BO{m B d^{4.5}}$ operations to solve the LP.
Thus, the computational complexities for the first, second, third steps are
 $\BO{|\mathcal{C}|^2 d^{4.5} B}$, $\BO{|\action| |\mathcal{C}| d^{4.5} B}$ and $\BO{|\action| |\mathcal{C}| d^{4.5} B}$, respectively.
 Hence, the computational complexity of OCP is
 $\BO{ \left[ |\action|+|\mathcal{C}| \right] |\mathcal{C}| d^{4.5} B}$ operations per period. \textbf{q.e.d.}\\
\endproof

Notice that the computational complexity is polynomial in
$d$, $B$, $|\mathcal{C}|$ and $|\mathcal{A}|$, and thus, OCP will be computationally efficient if all these parameters are tractably small.
Note that the bound in Proposition \ref{the:computational_complexity} is a worst-case bound, and the $O (d^{4.5})$ term is incurred by the need to solve LPs.
For some special cases, the computational complexity is much less. For instance, in the state aggregation case, the computational complexity is
$\BO{|\mathcal{C}|+|\mathcal{A}|+d}$ operations per period.

As we have discussed above,
one can ensure that $|\mathcal{C}|$ remains bounded by 
using variants of the constraint selection algorithm (Algorithm \ref{alg:ConstraintSelection})
that only use a subset of the available constraints.
For instance, in the coherent learning case discussed in Section \ref{sec:coherent_hypothesis_class}, we can use a
constraint selection algorithm that only chooses the constraints that will lead to a strict reduction of the eluder dimension
of the hypothesis class. 
Obviously, 
with this constraint selection algorithm, 
$|\mathcal{C}| \leq |\mathcal{C}_{-1}| + \mathrm{dim}_E \left( \mathcal{Q}\right) $ always holds, 
where $\mathcal{C}_{-1}$ is the set of constraints defining $\mathcal{Q}$. 
Similarly, in the state aggregation case considered in Section \ref{sec:agnostic_learning}, we can use a constraint selection algorithm 
that only chooses the constraints that reduce the optimistic Q-values of disjoint subsets from 
from infinity to finite.
Obviously, with this constraint selection algorithm,
$|\mathcal{C}| \leq |\mathcal{C}_{-1}| + K $ always holds, where 
$K$ is the number of partitions.
Based on our analysis, it can be shown that with these constraint selection algorithms, 
the performance bounds derived in Section \ref{sec:sample_efficiency} will still hold.
Finally, for the general agnostic learning case, 
one naive approach is to maintain a time window $W$, 
and only constraints observed in episode
$j-W, \cdots, j-1$ are used to construct $\mathcal{Q}_{\mathcal{C}}$ in episode $j$.

\section{Computational Results} \label{sec:experiment}
In this section, we present computational results involving two illustrative examples: 
the system presented in Example \ref{example:chain}
and the inverted pendulum problem considered in \citet{lagoudakis2002least}.  We 
compare OCP against least-squares value iteration (LSVI), a classical reinforcement learning algorithm.

\subsection{Learning in a Deterministic Chain}
Consider Example~\ref{example:chain} discussed in Section~\ref{sec:inefficient}.
Let $\phi_{t,k}$ be a feature mapping $\state \times \action$ to $\Re$ for any
$t=0,1,\cdots, H-1$ and any $k=1,2,\cdots, K$. We choose 
$\Qspace_t = \mathrm{span} \left (\phi_{t,1},  \ldots, \phi_{t,K} \right)$ and
$\Qspace = \Qspace_0 \times \cdots \times \Qspace_{H-1}$, and consider the coherent learning case with
$Q^* \in \Qspace$. 
Notice that when LSVI with Boltzmann/$\epsilon$-greedy exploration is applied to
this problem, the estimates for each period-state-action value $Q_t^*(x,a)$ will be $0$ until node
$N-1$ is first visited. 
Thus, as we have discussed in Section~\ref{sec:inefficient}, in expectation it will take 
the agent $2^{|\state|-1}$ episodes to first reach node $N-1$. Moreover, the lower bounds on $\reg(T)$ specified by Equation~\ref{eqn:example_1_regret_lb_1} and \ref{eqn:example_1_regret_lb_2} hold for any choice of $K$ and
any choice of features.


In our computational experiment, we choose $N=|\state|=H=50$, and 
simulate for 
$75000$ time steps (i.e., $1500$ episodes).
Obviously, for this choice of $N$ and $T$, $\reg(75000)$ 
can not exceed $1500$.
Based on the discussion above, if we apply LSVI with Boltzmann/$\epsilon$-greedy exploration to this problem,
for any choice of features,
in expectation it will take the agent $5.63 \times 10^{14}$ episodes ($2.81 \times 10^{16}$ time steps) to first reach node $N-1$,
and $\reg(75000) \geq 1500 - 3\times 10^{-12}$, which is extremely close to the worst-case regret. This shows that
LSVI with Boltzmann/$\epsilon$-greedy exploration is highly inefficient in this case.

\begin{figure}
\centering
\subfigure[$\reg$ vs. $T$ for $K=20$]
{
\label{fig:result_1}
\includegraphics[scale=0.48]{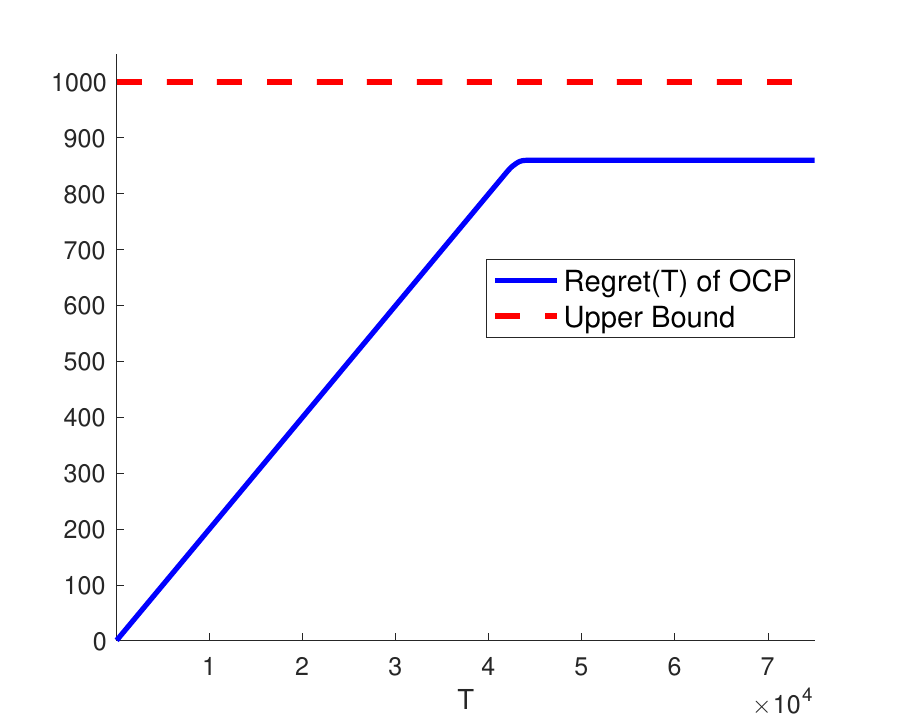}
}
\hspace{0.2in}
\subfigure[$\reg (75000)$ vs. $K$]
{
\label{fig:result_2}
\includegraphics[scale=0.48]{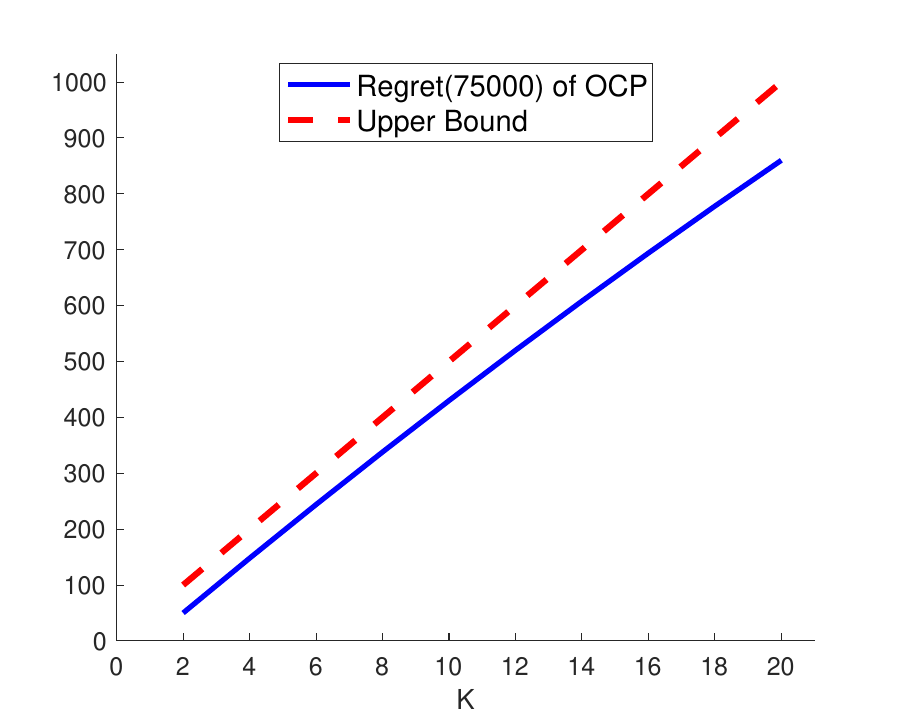}
}

\caption{Computational results for Example \ref{example:chain}}
\label{fig:experiment_result}
\end{figure}

We now describe our experiment setup for OCP, for which we need to specify how to choose features.
We are interested in how the performance of OCP scales with $K$, the number of features at each period $t$, and
vary $K=2, 4, 6, \cdots, 20$.
For a given $K$, we construct the features as follows: for each period $t=0,1,\cdots, H-1$, we choose $\phi_{t,1}=Q^*_t$,
$\phi_{t,2}=\mathbf{1}$, a vector of all ones, and if $K>2$, we sample $\phi_{t,3}, \cdots, \phi_{t,K} $ i.i.d. from the Gaussian distribution $N(0, I)$.
Notice that this ensures that $\Qspace$ is a coherent hypothesis class.
For $K=2$, we apply OCP to Example \ref{example:chain} with the above-specified features.
Notice that in this case, one simulation is sufficient since the features, the OCP algorithm, and the dynamic system are all deterministic.
On the other hand, for $K>2$, we apply OCP to Example \ref{example:chain} over $100$ repetitions, each time resampling features $\phi_{t,3}, \cdots, \phi_{t,K} $ for all $t$.
We then average the results of these $100$ simulations. 

Results are presented in Figure~\ref{fig:experiment_result}.
Specifically, in Figure~\ref{fig:result_1}, we fix $K=20$ and vary $T=50, 100, \cdots, 75000$,
and plot $\reg(T)$ as a function of $T$.
In Figure~\ref{fig:result_2}, we fix $T=75000$  and vary $K=2,4, \cdots, 20$, and plot
$\reg(75000)$ as a function of $K$. From Theorem~\ref{the:sample_efficiency}, in this problem,
the $O(1)$ bound on $\reg(T)$ of OCP is $HK=50K$. 
We also plot this $O(1)$ upper bound in the figures.

We now briefly discuss the results. Note that in this problem, the \emph{realized regret}
in an episode is either $0$ or $1$, 
depending on whether or not the agent reaches node $N-1$ in that episode (see Figure~\ref{fig:example1}).
Figure~\ref{fig:result_1} shows that for $K=20$, it takes the agent about $900$ episodes to learn how to reach node $N-1$.
Based on our discussion above, this result demonstrates the dramatic efficiency gains of OCP over LSVI with
Boltzmann/$\epsilon$-greedy exploration in this problem.
On the other hand, Figure~\ref{fig:result_2} shows that $\reg(75000)$ scales linearly with $K$.
The results also indicate that the $O(1)$ upper bound derived in Theorem~\ref{the:sample_efficiency} is not tight in this problem, 
but the gap is small.

\subsection{Inverted Pendulum}
We will now show that OCP significantly outperforms LSVI with $\epsilon$-greedy exploration in a reinforcement learning formulation 
of an inverted pendulum problem. The system dynamics of an inverted pendulum on a cart are described in Equation (18) of \citet{wang1996approach},
which is
\begin{align}
\dot{x}_1 =& \, x_2 \nonumber \\
\dot{x}_2 =& \, \frac{g \sin(x_1)-\alpha m l x_2^2 \sin(2x_1)/2-\alpha \cos(x_1) u}{4l/3-\alpha ml \cos^2 (x_1)}
\end{align}
where $x_1$ is the angular position (in radians) of the pendulum from the vertical, $x_2$ is the angular velocity,  
$g=9.8 \mathrm{m/s^2}$ is the gravity constant, $m=2\mathrm{kg}$ is the mass of the pendulum, $M=8\mathrm{kg}$ is the mass of the cart,
$l=0.5\mathrm{m}$ is the length of the pendulum, $\alpha=1/(m+M)=0.1\mathrm{kg}^{-1}$, and $u$ is the force applied to the cart (in Newtons).
Note that $\dot{x}_1$ and $\dot{x}_2$ are respectively the derivatives of $x_1$ and $x_2$ with respect to time.
Similarly as \citet{lagoudakis2002least}, we simulate this nonlinear system with a step size $0.1\mathrm{s}$. 
The action space $\action=\{-50, 0, 50 \}$, but the actual input to the system can be noisy.
Specifically, when action $a \in \action$ is selected, the actual input to the system is
$u=a+\xi_a$, where $\xi_a$ is a random variable independently drawn from the uniform distribution
$\mathrm{unif}(-\delta, \delta)$ for some $\delta \geq 0$. The initial state of the system is $(x_1=0, x_2=\xi_0)$,
where $\xi_0$ is also independently drawn from $\mathrm{unif}(-\delta, \delta)$.
Notice that this dynamic system is deterministic if $\delta=0$.

We consider a reinforcement learning setting in which an agent learns to control the inverted pendulum such that it does not fall for one hour while repeatedly interacting with it for $1000$ episodes. The reward in each episode $j$ is the length of time until the inverted pendulum falls, capped at one hour. We also assume that the agent does not know the system dynamics or the reward function. We apply OCP and LSVI with the same form of state aggregation to this problem. In particular, the state space of this problem is
\[ \state=\{(x_1, x_2) : \, x_1 \in (-\pi/2, \pi/2), \, x_2 \in \Re \} \bigcup \, \{ \text{inverted pendulum is fallen} \} .\]
We grid the angular position space
$ ( -\pi/2, \pi/2 )$ uniformly into $31$ intervals; and grid the angular velocity space as $(-\infty, -x_2^{\mathrm{max}})$, $(x_2^{\mathrm{max}}, \infty)$ and $29$ uniform intervals between $-x_2^{\mathrm{max}}$ and $x_2^{\mathrm{max}}$, where
$x_2^{\mathrm{max}}$ is the maximum angular velocity observed when the initial state is $(0,0)$ and $u=50$ for all the time steps.
$\state$ is partitioned as follows: the first partition only includes the special state ``inverted pendulum is fallen",
and all the other $961$ partitions are Cartesian products of intervals of $x_1$ and $x_2$ described above.
We choose the basis functions as the indicator functions for each action-(state space partition) pair,\footnote{The inverted pendulum problem is time-homogenous if it is not stopped by the time one hour. This motivates us to use basis functions independent of the period $t$.} 
hence there are $2886$ basis functions.

\begin{figure}
\centering
\subfigure[ $\delta=0$: LSVI vs Purely Randomized]
{
\label{fig:case1_LSVI}
\includegraphics[scale=0.38]{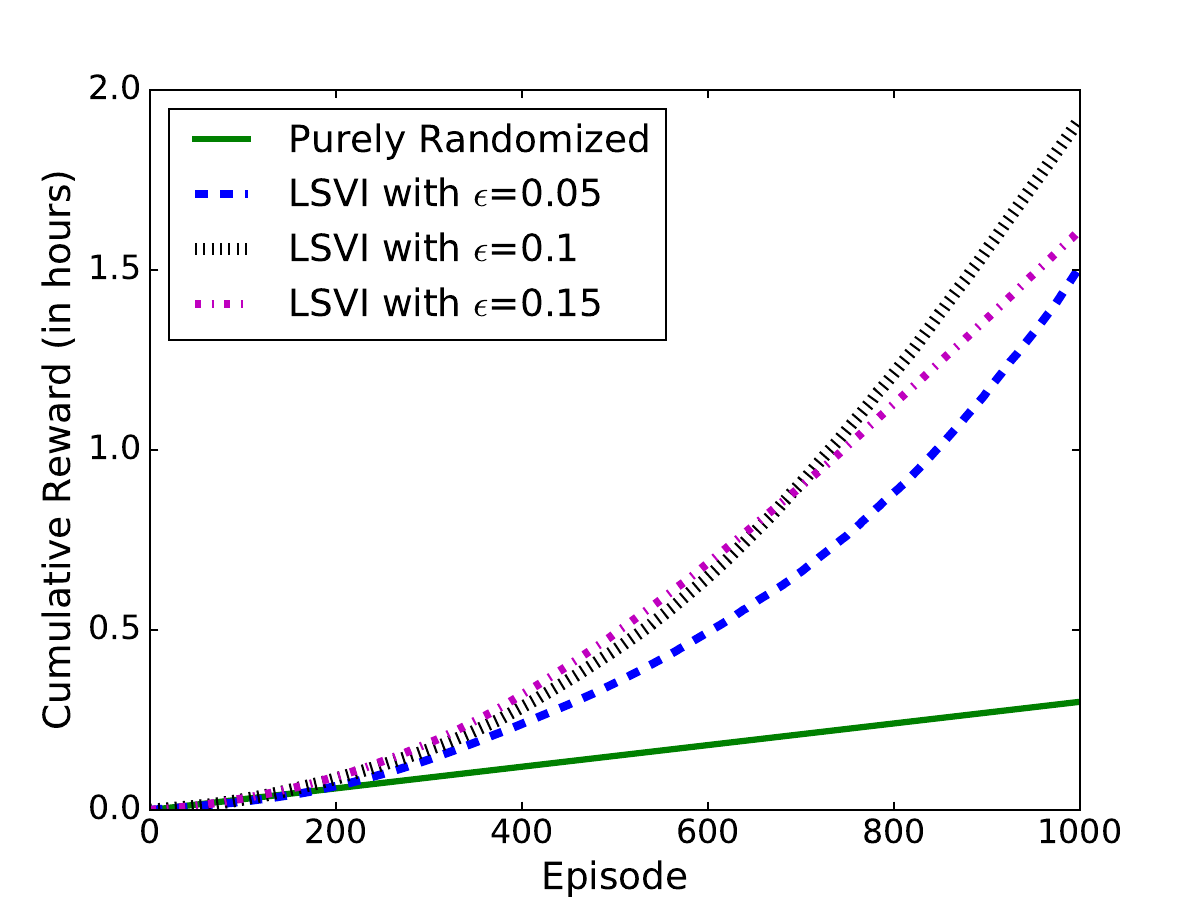}
}
\hspace{0.1in}
\subfigure[ $\delta=0$: OCP vs Best LSVI]
{
\label{fig:case1_OCP}
\includegraphics[scale=0.38]{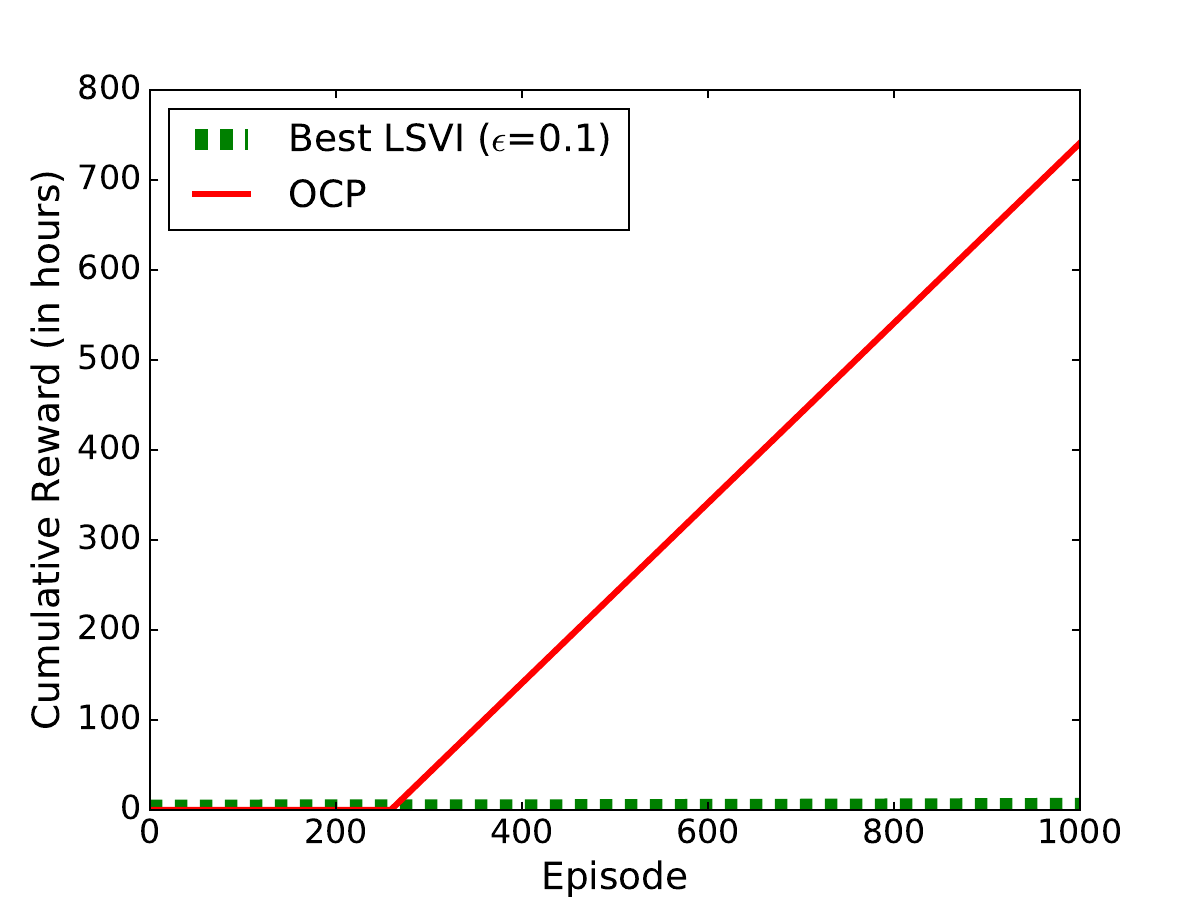}
}

\subfigure[ $\delta=2.5$: LSVI vs Purely Randomized]
{
\label{fig:case2_LSVI}
\includegraphics[scale=0.38]{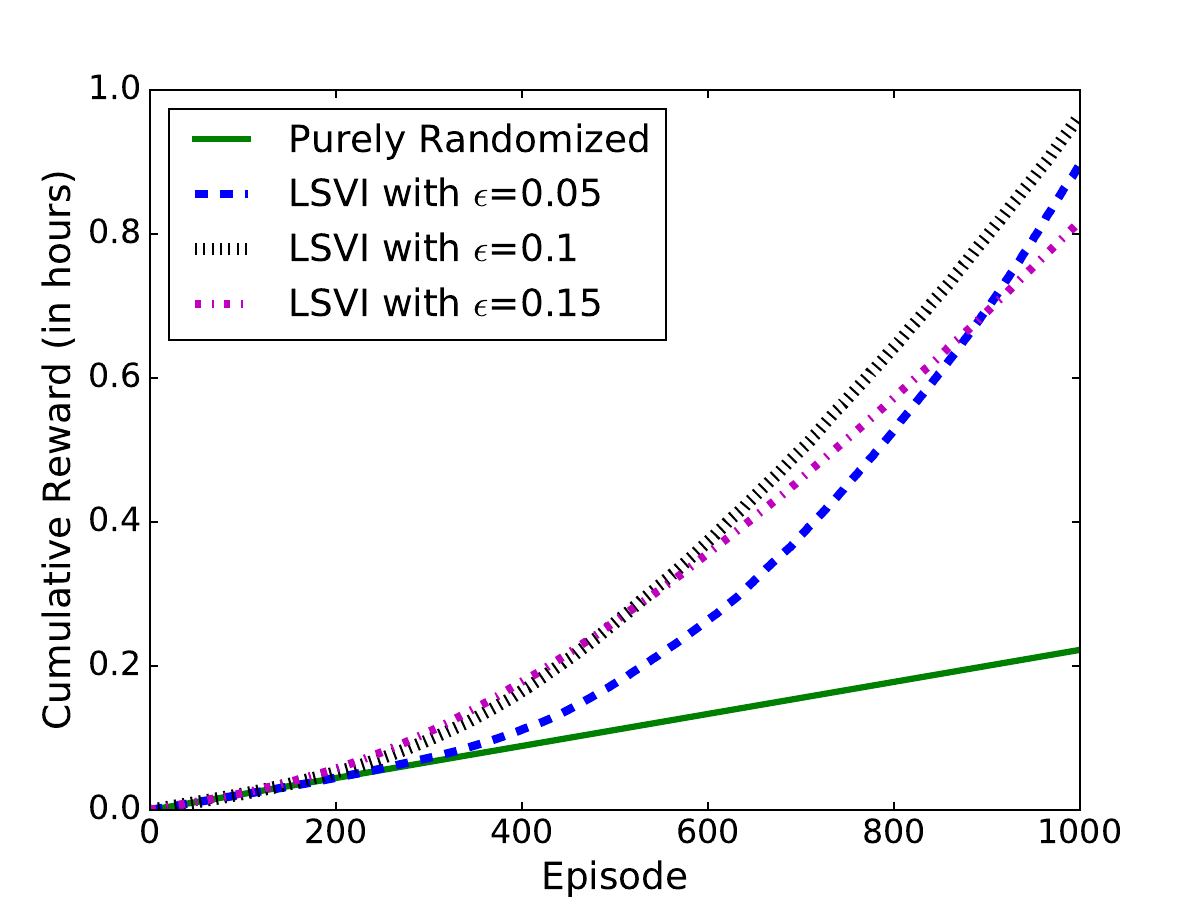}
}
\hspace{0.1in}
\subfigure[ $\delta=2.5$: OCP vs Best LSVI]
{
\label{fig:case2_OCP}
\includegraphics[scale=0.38]{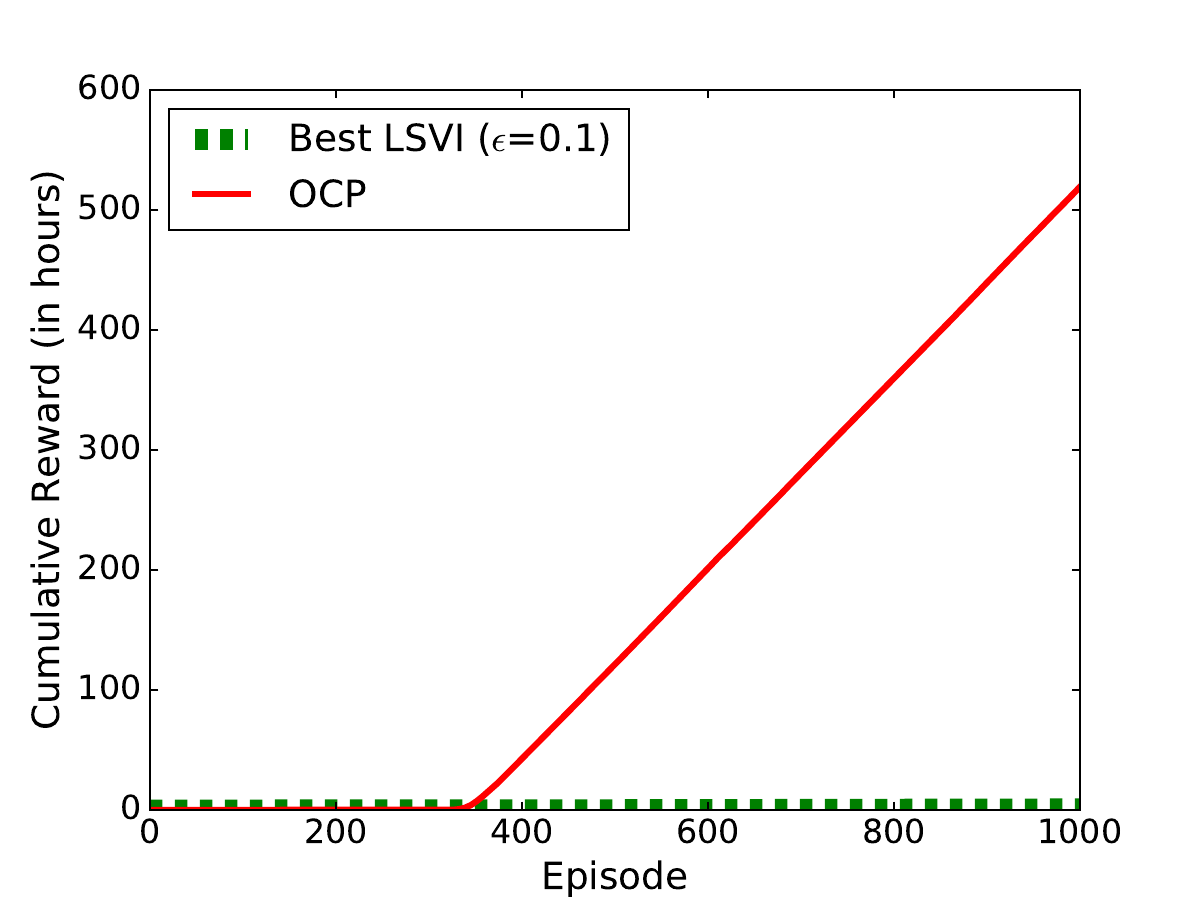}
}

\caption{Computational results for the inverted pendulum}
\label{fig:inverted_pendulum}
\end{figure}

We present computational results for two cases: $\delta=0$ and $\delta=2.5$. For each case, we apply OCP and LSVI with exploration rate $\epsilon=0.05, 0.1, 0.15$ to it. We also show the performance of a purely randomized policy as a baseline, under which each action in $\action$ is chosen uniformly randomly at each time. Results are averaged over $100$ simulations. Figure~\ref{fig:inverted_pendulum} plots the cumulative reward as a function of episode.
Notice that the cumulative reward in the first $J$ episodes is bounded by $J$ hours since the per-episode reward is upper bounded by one hour.

Figure~\ref{fig:case1_LSVI} and \ref{fig:case2_LSVI} compare LSVI with $\epsilon$-greedy exploration with the purely randomized policy. Notice that though LSVI significantly outperforms the purely randomized policy, its performance is unsatisfactory since in both cases its cumulative reward at $1000$ episodes is less than $2$ hours, indicating that in the first $1000$ episodes the average time length until the pendulum falls is less than $7.2$ seconds. Figure~\ref{fig:case1_OCP} and \ref{fig:case2_OCP} compare OCP with the best LSVI ($\epsilon=0.1$ in both cases). We observe that in both cases, the performance of OCP is orders of magnitude better than that of the LSVI.
We also note that the performances of both OCP and LSVI are worse in the case with $\delta=2.5$ than the case with $\delta=0$, since the stochastic disturbances make the inverted pendulum problem more challenging.

Finally, we would like to emphasize that the system dynamics are stochastic in the case with $\delta=2.5$. 
However, the magnitude of the stochastic disturbances, $2.5$, is small relative to the magnitude of the control,
$50$.
Thus, though OCP is motivated and developed in the framework of reinforcement learning in deterministic systems, 
it might also perform well in some reinforcement learning problems with stochastic environments (e.g. reinforcement learning in MDPs),
especially when the magnitude of the stochastic disturbances is small.

\section{Conclusion} \label{sec:conclusion}
We have proposed a novel reinforcement learning algorithm, called optimistic constraint propagation (OCP), 
that synthesizes efficient exploration and value function generalization for episodic reinforcement learning in deterministic systems.
We have shown that when the true value function $Q^*$ lies in the given hypothesis class $\Qspace$ (the coherent learning case),
OCP selects optimal actions over all but at most $\dimM{\Qspace}$ episodes, where
$\dimM{\Qspace}$ is the eluder dimension of $\Qspace$.
We have also established sample efficiency and asymptotic performance guarantees for the state aggregation case,
a special agnostic learning case where $\Qspace$ is the span of pre-specified indicator functions over disjoint sets. 
We have also discussed the computational complexity of OCP and presented computational results involving two
illustrative examples.
Our results demonstrate dramatic efficiency gains enjoyed by OCP relative to LSVI with 
Boltzmann or $\epsilon$-greedy exploration.

Finally, we briefly discuss some possible directions for future research. 
One possible direction is to propose a variant of OCP for reinforcement learning in infinite-horizon discounted deterministic systems.
Note that for an infinite-horizon discounted problem with bounded rewards, its \emph{effective horizon} is
$\frac{1}{1-\gamma}$, where $\gamma \in (0,1)$ is the discount factor.
We conjecture that with this notion of effective horizon,
similar 
sample complexity/regret bounds can be derived for the infinite-horizon discounted problems.
Another possible direction is to design provably  
sample efficient algorithms for the general agnostic learning case discussed in this paper.
A more important problem is to design efficient algorithms for reinforcement learning in MDPs. 
Though many provably efficient algorithms have been proposed for the {\it tabula rasa} case of this problem (see \citep{Brafman2002, Strehl2006, neu2012adversarial, osband2013more, vanroy2014generalization} and references therein), however,
how to design such algorithms with value function generalization is currently still open. Thus, one interesting
direction for future research is to extend OCP, or a variant of it, to this problem.

\begin{APPENDICES}

\section{Equivalence of OCP and Q-Learning in the Tabula Rasa Case}

We prove that in the finite state/action {\it tabula rasa} case, 
OCP is equivalent to Q-learning with learning rate $1$ and initial Q-value $Q_t(x,a) =\infty$.
To see it, notice that in this setting, the OCP algorithm imposes constraints on individual Q-values of all the state-action-period triples.
Moreover, if we define the optimistic Q-function in an arbitrary episode $j$ as
\[
\Qopt_t (x,a) = \sup_{Q \in \Qspace_{\CC}} Q_t (x,a) \quad \forall (x,a,t),
\]
then $a_{j,t} \in \argmax_{a \in \action} \Qopt_t (x_{j,t},a)$. 
Thus, the lower bound $L_{j,t}$'s do not matter in this setting 
since there is no value function generalization across $(x,a,t)$'s.

Note that by definition of $\Qopt$, $U_{j,t} = R_t (x_{j,t}, a_{j,t}) + \sup_{a \in \action} \Qopt_{t+1} (x_{j,t+1}, a)$.
Moreover, since $Q^* \in \Qspace$ in this case, as we will prove in Lemma \ref{lemma:tech0}, there are no conflicting constraints.
Hence, in the next episode (episode $j+1$), the optimistic Q-function is updated as
\[
\Qopt_t (x,a) \leftarrow \left \{
\begin{array}{ll}
\min \left \{ \Qopt_t (x,a) , U_{j,t} \right \} & \text{if $(x,a,t)=(x_{j,t}, a_{j,t}, t)$} \\
\Qopt_t (x,a) & \text{otherwise} 
\end{array}
\right.
\]
Notice that the above equation implies that $\Qopt$ is a non-increasing function in episode
$j$.
Thus, to show OCP is equivalent to Q-learning, we only need to prove that
$U_{j,t} \leq \Qopt_t (x_{j,t}, a_{j,t})$ in episode $j$.
Obviously, we only need to consider the case when $\Qopt_t (x_{j,t}, a_{j,t}) < \infty$.
Notice that this holds trivially for $t=H-1$, since when $t=H-1$,
$U_{j,t}=R_t (x_{j,t}, a_{j,t})$ always holds, and if $\Qopt_t (x_{j,t}, a_{j,t}) < \infty$
then $\Qopt_t (x_{j,t}, a_{j,t}) = R_t (x_{j,t}, a_{j,t})$.
On the other hand, if $t<H-1$ and $\Qopt_t (x_{j,t}, a_{j,t}) < \infty$, then 
$\Qopt_t (x_{j,t}, a_{j,t}) = U_{j',t}$ for some $j'<j$.
Note that by definition of $U_{j,t}$, we have
$\Qopt_t (x_{j,t}, a_{j,t}) = U_{j',t} \geq U_{j,t}$ since $\Qopt_{t+1}$ is a non-increasing function in
$j$.

\section{Eluder Dimension for the Sparse Linear Case}
We start by defining some useful terminologies and notations.
For any $\theta \in \Re^K$, any $l \le K$ and any index set $\mathcal{I}= \left \lbrace i_1, i_2, \cdots, i_l
\right \rbrace \subseteq \left \lbrace 1,2,\cdots, K \right \rbrace$ with
$i_1 < i_2 < \cdots < i_l$ and $|\mathcal{I}|=l \le K$,
we use $\theta_{\mathcal{I}}$ to denote the subvector of $\theta$
associated with the index set $\mathcal{I}$, i.e.
$
\theta_{\mathcal{I}} = \left[
\theta_{i_1}, \theta_{i_2} \cdots, 
\theta_{i_l}
\right]^T$.

For a sequence of vectors $\theta^{(1)}, \theta^{(2)}, \cdots \in \Re^K $, we say
$\theta^{(k)}$ is linearly $l$-independent of its predecessors if 
there exists an index set $\mathcal{I}$ with $|\mathcal{I}|=l$ s.t. 
$\theta^{(k)}_{\mathcal{I}}$ is linearly independent of 
$\theta^{(1)}_{\mathcal{I}}, \theta^{(2)}_{\mathcal{I}}, \cdots, \theta^{(k-1)}_{\mathcal{I}}$.
Let $N=|\state||\action|$, and use $\Phi_j^T$ to denote the $j$th row of $\Phi$.
For any $l \le K$, we define
$\rank [\Phi, l]$, the $l$-rank of $\Phi$,
as the length $d$ of the longest sequence of $\Phi_j$'s such that every element is linearly
$l$-independent of its predecessors.
Recall that $\Qspace_0 = \left \{ \Phi \theta: \, \theta \in \Re^K, \| \theta \|_0 \le K_0 \right \}$, we have the following result:
\begin{proposition}
If $2K_0 \le K$, then
$\dimM{\Qspace_0}=\rank[\Phi, 2K_0]$.
\label{p1}
\end{proposition}
\proof{Proof}
We use $y=(x,a)$ to denote a state-action pair, and use 
$\Phi(y)^T$ to denote the row of matrix $\Phi$ associated with $y$.
Based on our definitions of eluder dimension and $l$-rank, it is sufficient to prove the following lemma:

\begin{lemma}
For any state-action pair $y$ and
for any set of state-action pairs $Y= \left \lbrace y^{(1)}, y^{(2)}, \cdots,  y^{(n)} \right \rbrace$, $y$ is independent of $Y$ with respect to
$\Qspace_0$
if and only if
$\Phi(y)$ is linearly $2K_0$-independent of 
$\left \lbrace \Phi( y^{(1)}) , \Phi ( y^{(2)} ), \cdots, \Phi (y^{(n)}) \right \rbrace$.
\end{lemma}

\noindent
We now prove the above lemma.
Note that based on the definition of independence (see Section \ref{sec:eluder_dimension}), 
$y$ is independent of $Y$ with respect to
$\Qspace_0$
if and only if there exist $Q_1, Q_2 \in \Qspace_0$ s.t.
$Q_1 (y^{(i)})=Q_2 (y^{(i)})$,
$\forall i=1,2,\cdots, n$, and
$Q_1 (y) \neq Q_2 (y)$.
Based on the definition of function space $\Qspace_0$, there exist two $K_0$-sparse vectors
$\theta^{(1)}, \theta^{(2)} \in \Re^K$ s.t. $Q_1=\Phi \theta^{(1)}$ and 
$Q_2=\Phi \theta^{(2)}$.
Thus, 
$y$ is independent of $Y$ with respect to
$\Qspace_0$
if and only if
there exist two $K_0$-sparse vectors
$\theta^{(1)}, \theta^{(2)} \in \Re^K$ s.t.
\BE
\Phi(y^{(i)})^T (\theta^{(1)}-\theta^{(2)}) &=& 0 \quad \forall i=1,2,\cdots, n \nonumber \\
\Phi(y)^T (\theta^{(1)}-\theta^{(2)}) &\neq & 0 \nonumber
\EE
Based on the definition of $K_0$-sparsity, the above condition is 
equivalent to
there exists a $2K_0$-sparse vector $\theta \in \Re^K$
s.t.
\BE
\Phi(y^{(i)})^T \theta &=& 0 \quad \forall i=1,2,\cdots, n \nonumber \\
\Phi(y)^T \theta &\neq & 0 \nonumber
\EE
To see it, note that if $\theta^{(1)}, \theta^{(2)}$ are
$K_0$-sparse, then $\theta=\theta^{(1)}-\theta^{(2)}$ is $2K_0$-sparse.
On the other hand, if $\theta$ is $2K_0$-sparse, then there exist two 
$K_0$-sparse vectors $\theta^{(1)}, \theta^{(2)}$ s.t.
$\theta=\theta^{(1)}-\theta^{(2)}$.

Since $\theta$ is $2K_0$-sparse,
there exists a set of indices $\mathcal{I}$
s.t. $|\mathcal{I}|=2K_0$ and
$\theta_i=0$, $\forall i \notin \mathcal{I}$.
Thus, the above condition is equivalent to
\BE
\Phi(y^{(i)})_{\mathcal{I}}^T \theta_{\mathcal{I}} &=& 0 \quad \forall i=1,2,\cdots, n \nonumber \\
\Phi(y)_{\mathcal{I}}^T \theta_{\mathcal{I}} &\neq & 0, \nonumber
\EE
which is further equivalent to $\Phi(y)_{\mathcal{I}}$
is linearly independent of $\Phi(y^{(1)})_{\mathcal{I}},
\Phi(y^{(2)})_{\mathcal{I}}, \cdots, \Phi(y^{(n)})_{\mathcal{I}}$.
Since $|\mathcal{I}|=2K_0$,
from the definition of linear $l$-dependence, this is equivalent to 
$\Phi(y)$
is linearly $2K_0$-independent of $\Phi(y^{(1)}),
\Phi(y^{(2)}), \cdots, \Phi(y^{(n)})$. \textbf{q.e.d.}
\endproof

We now show that if $\Phi$ satisfies a technical condition, then
$\rank [\Phi,l]=l$. Specifically, for any $l \le \min \{ N,K \}$, we say $\Phi$ is $l$-full-rank if any submatrix
of $\Phi$ with size $l \times l$ has full rank.
Based on this notion, we have the following result:
\begin{proposition}
For any $l \le \min \{ N,K \}$,
if $\Phi$ is $l$-full-rank, then we have $\rank[\Phi, l]=l$. 
\label{p3}
\end{proposition}

\proof{Proof}
Consider any sequence of matrix rows $\Phi^{(1)}, \Phi^{(2)}, \cdots, \Phi^{(l+1)}$
with length $l+1$, and any index set $\mathcal{I}$ with $|\mathcal{I}|=l$.
Since $\Phi$ is $l$-full-rank, thus $\Phi^{(1)}_{\mathcal{I}}, \Phi^{(2)}_{\mathcal{I}}, \cdots, \Phi^{(l)}_{\mathcal{I}} \in \Re^l$
are linearly independent (hence forms a basis in $\Re^l$). Thus, $\Phi^{(l+1)}_{\mathcal{I}}$ is linearly dependent on 
$\Phi^{(1)}_{\mathcal{I}}, \Phi^{(2)}_{\mathcal{I}}, \cdots, \Phi^{(l)}_{\mathcal{I}} \in \Re^l$. Since this result holds for any $\mathcal{I}$
with $|\mathcal{I}|=l$, thus $\Phi^{(l+1)}$ is linearly $l$-dependent on $\Phi^{(1)}, \Phi^{(2)}, \cdots, \Phi^{(l)} \in \Re^K$.
Furthermore, since this result holds for any sequence of matrix rows with length $l+1$, thus we have
$\rank[\Phi, l]\le l$.

On the other hand, since $\Phi$ is $l$-full-rank, choose any sequence of matrix rows $\Phi^{(1)}, \Phi^{(2)}, \cdots, \Phi^{(l)}$ with length $l$ and
any index set $\mathcal{I}$ with $|\mathcal{I}|=l$, $\Phi^{(1)}_{\mathcal{I}}, \Phi^{(2)}_{\mathcal{I}}, \cdots, \Phi^{(l)}_{\mathcal{I}}$
are linearly independent. Thus, $\Phi^{(1)}, \Phi^{(2)}, \cdots, \Phi^{(l)}$ is a sequence of matrix rows s.t. every element is linearly $l$-independent of its predecessors.
Thus, $\rank[\Phi, l]\ge l$. So we have
$\rank[\Phi, l] = l$. \textbf{q.e.d.}
\endproof

Thus, if $2K_0 \le \min \{ N, K \}$ and $\Phi$ is $2K_0$-full-rank, then we have
$\dimM{\Qspace_0}=\rank \left[ \Phi, 2K_0 \right]=2K_0$.
Consequently, we have $\dimM{\Qspace}=\dimM{\Qspace_0^H}=2K_0 H$.

\section{Detailed Proof for Theorem \ref{the:sample_efficiency}}
\subsection{Proof for Lemma \ref{lemma:tech0}}
\proof{Proof for Lemma \ref{lemma:tech0}}

We prove this lemma by induction on $j$ and choose the induction hypothesis as follows:
$\forall j=0,1,\cdots$, we have
(1) $Q^* \in \mathcal{Q}_{\CC_j}$ and (2) 
$L_{j',t} \leq Q^*_t (x_{j',t}, a_{j',t}) \leq U_{j',t}$ for all $t=0,1,\cdots, H-1$ and all $j'=0,1,\cdots, j-1$.

First, we notice that the induction hypothesis is true for $j=0$. To see it, notice that when $j=0$, (2) holds trivially since $j-1<0$; and
(1) also holds since by definition $\mathcal{Q}_{\CC_0}=\Qspace$, and hence $Q^* \in \Qspace = \mathcal{Q}_{\CC_0}$.
We now prove that if the induction hypothesis holds for episode $j$, then it also holds for episode $j+1$.
We first show that (2) holds for episode $j+1$, which is sufficient to prove 
\[
L_{j,t} \leq Q^*_t (x_{j,t}, a_{j,t}) \leq U_{j,t} \quad \forall t=0,1,\cdots, H-1.
\]
We prove the above inequality by considering two different cases. First, if $t=H-1$, then we have
$U_{j,t}=L_{j,t}=R_{t} (x_{j,t}, a_{j,t}) = Q^*_t (x_{j,t}, a_{j,t})$, and hence the above inequality trivially holds.
On the other hand, if $t<H-1$, then we have
\BE
U_{j,t} &=& R_{t} (x_{j,t}, a_{j,t}) + \sup_{Q \in \mathcal{Q}_{\CC_j}} \sup_{a \in \action} Q_{t+1} (x_{j,t+1}, a) \nonumber \\
& \geq &  R_{t} (x_{j,t}, a_{j,t}) + \sup_{a \in \action} Q^*_{t+1} (x_{j,t+1}, a) = Q^*_{t} (x_{j,t}, a_{j,t}), \nonumber
\EE
where the inequality follows from the induction hypothesis $Q^* \in \mathcal{Q}_{\CC_j}$, and the last equality follows from the Bellman equation.
Similarly, we also have
\BE
L_{j,t} &=& R_{t} (x_{j,t}, a_{j,t}) + \inf_{Q \in \mathcal{Q}_{\CC_j}} \sup_{a \in \action} Q_{t+1} (x_{j,t+1}, a) \nonumber \\
& \leq &  R_{t} (x_{j,t}, a_{j,t}) + \sup_{a \in \action} Q^*_{t+1} (x_{j,t+1}, a) = Q^*_{t} (x_{j,t}, a_{j,t}). \nonumber
\EE
Hence, (2) holds for episode $j+1$. Since $Q^* \in \Qspace$ and (2) holds for episode $j+1$, then by definition of $\mathcal{Q}_{\CC_{j+1}}$,
we have $Q^* \in \mathcal{Q}_{\CC_{j+1}}$. 
Thus, the induction hypothesis also holds for episode $j+1$.
Hence, we have completed the proof for Lemma \ref{lemma:tech0}.
\textbf{q.e.d.}
\endproof
\subsection{Proof for Lemma \ref{lemma:tech1}}
\proof{Proof for Lemma \ref{lemma:tech1}}
We prove this lemma by induction on $j$. First, notice that if $j=0$, then from Algorithm \ref{alg:auxiliary}, we have
$\SAT_0=\varnothing$. Thus, Lemma \ref{lemma:tech1}(a) holds for $j=0$.

Second, we prove that if Lemma \ref{lemma:tech1}(a) holds for episode $j$, then Lemma \ref{lemma:tech1}(b) holds for episode $j$
and Lemma \ref{lemma:tech1}(a) holds for episode $j+1$.
To see why Lemma \ref{lemma:tech1}(b) holds for episode $j$, notice that from Lemma \ref{lemma:tech0}, we have $Q^* \in \Qspace_{\CC_{j}} \subseteq \Qspace$. 
Furthermore, from the induction hypothesis, $\forall \sat \in \SAT_j$ and $\forall Q \in \Qspace_{\CC_{j}}$, we have
$Q (\sat)=Q^* (\sat)$.
Since $(x_{j,t}, a_{j,t}, t)$ is dependent on $\SAT_j$ with respect to $\Qspace$, then $ \forall Q \in \Qspace_{\CC_{j}} \subseteq \Qspace$,
we have that $Q_t (x_{j,t}, a_{j,t}) = Q^*_t (x_{j,t}, a_{j,t})$.
Hence we have $\sup_{Q \in \Qspace_{\CC_{j}}} Q_t(x_{j,t}, a_{j,t})=Q^*_t (x_{j,t}, a_{j,t})$, furthermore, from the OCP algorithm, we 
have $\sup_{Q \in \Qspace_{\CC_{j}}} Q_t(x_{j,t}, a_{j,t}) \ge \sup_{Q \in \Qspace_{\CC_{j}}} Q_t(x_{j,t}, a)$, $\forall a \in \action$, thus we
have
\[
Q^*_t (x_{j,t}, a_{j,t})=\sup_{Q \in \Qspace_{\CC_{j}}} Q_t(x_{j,t}, a_{j,t}) \ge \sup_{Q \in \Qspace_{\CC_{j}}} Q_t(x_{j,t}, a) \ge 
Q^*_t(x_{j}, a), \quad \forall a \in \action,
\]
where the last inequality follows from the fact that $Q^* \in \Qspace_{\CC_{j}}$. Thus, $a_{j,t}$ is optimal and
$Q^*_t (x_{j,t}, a_{j,t})=V^*_t (x_{j,t})$.
Thus, Lemma \ref{lemma:tech1}(b) holds for episode $j$.

We now prove Lemma \ref{lemma:tech1}(a) holds for episode $j+1$. We prove the conclusion by considering two different scenarios. If 
$t_j^*=\NULL$, then $\SAT_{j+1}=\SAT_j$ and $\Qspace_{\CC_{j+1}} \subseteq \Qspace_{\CC_{j}}$. Thus, obviously, Lemma \ref{lemma:tech1}(a) holds for episode $j+1$.
On the other hand, if $t_j^* \neq \NULL$, we have $\Qspace_{\CC_{j+1}} \subseteq \Qspace_{\CC_{j}}$ and
$\SAT_{j+1} = \left[ \SAT_j ,  (x_{j,t_j^*}, a_{j,t_j^*}, t_j^*) \right]$.
Based on the induction hypothesis, $\forall \sat \in \SAT_j$ and $\forall Q \in \Qspace_{\CC_{j+1}} \subseteq \Qspace_{\CC_{j}}$, we have
$Q(\sat)=Q^*(\sat)$.
Thus, it is sufficient to prove that 
\BE
Q_{t_j^*} (x_{j,t_j^*}, a_{j,t_j^*})=Q^*_{t_j^*} (x_{j,t_j^*}, a_{j,t_j^*}), \quad \forall Q \in \Qspace_{\CC_{j+1}}.  \label{lemma:tech1:eq2}
\EE
We prove Eqn(\ref{lemma:tech1:eq2}) by considering two different cases. First, if $t_j^*=H-1$, it is sufficient to prove that
$Q_{H-1} (x_{j,H-1}, a_{j,H-1})=R_{H-1} (x_{j,H-1}, a_{j,H-1})$, $\forall Q \in \Qspace_{\CC_{j+1}}$,
which holds by definition of $\Qspace_{\CC_{j+1}}$ (see OCP algorithm, and recall that from Lemma \ref{lemma:tech0}, no constraints are conflicting if $Q^* \in \Qspace$).
On the other hand, if $t_j^*<H-1$, it is sufficient to prove that for any $Q \in \Qspace_{\CC_{j+1}}$,
$Q_{t_j^*} (x_{j,t_j^*}, a_{j,t_j^*})=R_{t_j^*} (x_{j,t_j^*}, a_{j,t_j^*})+ V^*_{t_j^*+1} (x_{j,t_j^*+1})$.
Recall that OCP algorithm add a constraint 
$L_{j,t_j^*} \leq Q_{t_j^*} (x_{j,t_j^*}, a_{j,t_j^*}) \leq U_{j,t_j^*}$
to $\Qspace_{\CC_{j+1}}$ (and again, recall that no constraints are conflicting if $Q^* \in \Qspace$).
Based on the definitions of $L_{j,t_j^*}$ and $U_{j,t_j^*}$,
it is sufficient to prove that
\BE
V^*_{t_j^*+1} (x_{j,t_j^*+1})=\sup_{Q \in \Qspace_{\CC_{j}}} \sup_{a \in \action} Q_{t_j^*+1} (x_{j,t_j^*+1}, a) 
= \inf_{Q \in \Qspace_{\CC_{j}}} \sup_{a \in \action} Q_{t_j^*+1} (x_{j,t_j^*+1}, a). \label{lemma:tech1:eq4}
\EE
We first prove that
$V^*_{t_j^*+1} (x_{j,t_j^*+1})=\sup_{Q \in \Qspace_{\CC_{j}}} \sup_{a \in \action} Q_{t_j^*+1} (x_{j,t_j^*+1}, a)$. 
Specifically, we have that
\[
\sup_{Q \in \Qspace_{\CC_{j}}} \sup_{a \in \action} Q_{t_j^*+1} (x_{j,t_j^*+1}, a)=
\sup_{a \in \action} \sup_{Q \in \Qspace_{\CC_{j}}} Q_{t_j^*+1} (x_{j,t_j^*+1}, a) =
\sup_{Q \in \Qspace_{\CC_{j}}} Q_{t_j^*+1} (x_{j,t_j^*+1}, a_{j,t_j^*+1})=
 V^*_{t_j^*+1} (x_{j,t_j^*+1}),
\]
where
the second equality follows from the fact that
 $a_{j,t_j^*+1} \in \argmax_{a \in \action} \sup_{Q \in \Qspace_{\CC_{j}}} Q_{t_j^*+1} (x_{j,t_j^*+1}, a)$ and
the last equality follows from
the definition of $t_j^*$ and Part (b) of the lemma for episode $j$ (which we have just proved above, and holds by the induction hypothesis). Specifically, since $t_j^*$ is the last period in episode $j$ s.t. 
$(x_{j,t}, a_{j,t}, t)$ is independent of $\SAT_j$ with respect to $\Qspace$. Thus,
$(x_{j,t_j^*+1}, a_{j,t_j^*+1}, t_j^*+1)$ is dependent on $\SAT_j$ with respect to $\Qspace$. From Lemma \ref{lemma:tech1}(b) for episode $j$, we have
$V^*_{t_j^*+1} (x_{j,t_j^*+1})=Q_{t_j^*+1} (x_{j,t_j^*+1}, a_{j,t_j^*+1})$ for any
$Q \in \Qspace_{\CC_{j}}$. Thus, 
$\sup_{Q \in \Qspace_{\CC_{j}}} Q_{t_j^*+1} (x_{j,t_j^*+1}, a_{j,t_j^*+1})=
 V^*_{t_j^*+1} (x_{j,t_j^*+1})=\inf_{Q \in \Qspace_{\CC_{j}}} Q_{t_j^*+1} (x_{j,t_j^*+1}, a_{j,t_j^*+1})$.
On the other hand, we have that
\[
\inf_{Q \in \Qspace_{\CC_{j}}} \sup_{a \in \action} Q_{t_j^*+1} (x_{j,t_j^*+1}, a) \geq
 \sup_{a \in \action} \inf_{Q \in \Qspace_{\CC_{j}}}  Q_{t_j^*+1} (x_{j,t_j^*+1}, a) \geq
 \inf_{Q \in \Qspace_{\CC_{j}}}  Q_{t_j^*+1} (x_{j,t_j^*+1}, a_{j,t_j^*+1})=V^*_{t_j^*+1} (x_{j,t_j^*+1}),
\]
where the first inequality follows from the max-min inequality, the second inequality
follows from the fact that
$a_{j,t_j^*+1} \in \action$, and we have just proved the last equality above.
Hence we have
 \[
 V^*_{t_j^*+1} (x_{j,t_j^*+1})=\sup_{Q \in \Qspace_{\CC_{j}}} \sup_{a \in \action} Q_{t_j^*+1} (x_{j,t_j^*+1}, a) \geq 
 \inf_{Q \in \Qspace_{\CC_{j}}} \sup_{a \in \action} Q_{t_j^*+1} (x_{j,t_j^*+1}, a) \geq
 V^*_{t_j^*+1} (x_{j,t_j^*+1}). 
 \]
Thus, Eqn(\ref{lemma:tech1:eq4}) holds. 
Hence, Lemma \ref{lemma:tech1}(a) holds for episode $j+1$, and by induction, we have proved Lemma \ref{lemma:tech1}.
\textbf{q.e.d.}
\endproof

\subsection{Proof for Lemma \ref{lemma:ee1}}
\proof{Proof for Lemma \ref{lemma:ee1}}
Note that from Algorithm \ref{alg:auxiliary}, if $t_j^*=\NULL$, then $\forall t=0,1,\cdots,H-1$, 
$(x_{j,t}, a_{j,t}, t)$ is dependent on $\SAT_j$ with respect to $\Qspace$.
Thus, from Lemma \ref{lemma:tech1}(b), $a_{j,t}$ is optimal $\forall t=0,1,\cdots,H-1$.
Hence we have
$
R^{(j)}=\sum_{t=0}^{H-1} R_t (x_{j,t}, a_{j,t})=V^*_0 (x_{j,0})$.

On the other hand, $t_j^*\neq \NULL$, then from Algorithm \ref{alg:auxiliary},
$(x_{j,t_j^*}, a_{j,t_j^*}, t_j^*)$ is independent of $\SAT_j$ and $|\SAT_{j+1}|=|\SAT_j|+1$.
Note $(x_{j,t_j^*}, a_{j,t_j^*}, t_j^*) \in \SAT_{j+1}$, hence from Lemma \ref{lemma:tech1}(a),
$\forall Q \in \Qspace_{\CC_{j+1}}$, we have
$Q_{t_j^*} (x_{j,t_j^*}, a_{j,t_j^*})=Q^*_{t_j^*} (x_{j,t_j^*}, a_{j,t_j^*})$. \textbf{q.e.d.}
\endproof
\noindent

\subsection{Proof for Theorem \ref{the:sample_efficiency} Based on Lemma \ref{lemma:ee1}}
\proof{Proof for Theorem \ref{the:sample_efficiency}}
Notice that $\forall j=0,1,\cdots$,
$R^{(j)} \le V_0^* (x_{j,0})$  by definition. Thus, from Lemma \ref{lemma:ee1},
$R^{(j)} < V_0^* (x_{j,0})$ implies that $t_j^* \neq \NULL$. Hence, for any  
$ j=0,1,\cdots$, we have
$\indicator \left[ R^{(j)}< V^*_0 (x_{j,0}) \right] \le \indicator \left[ t_j^* \neq \NULL \right].$
Furthermore, notice that from the definition of $\SAT_j$, we have
$\indicator \left[ t_j^* \neq \NULL \right] = |\SAT_{j+1}|-|\SAT_j| $,
where $|\cdot|$ denotes the length of the given sequence.
Thus for any $J=0,1,\cdots$, we have
\BE
\sum_{j=0}^{J} \indicator \left[ R^{(j)}< V^*_0 (x_{j,0}) \right] \le \sum_{j=0}^{J} \indicator \left[ t_j^* \neq \NULL \right] =
\sum_{j=0}^{J} \left[ |\SAT_{j+1}|-|\SAT_j|  \right]=|\SAT_{J+1}|-|\SAT_0|=|\SAT_{J+1}|, \label{theorem1:1}
\EE
where the last equality follows from the fact that $|\SAT_0|=|\varnothing|=0$.
Notice that by definition (see Algorithm \ref{alg:auxiliary}),  $\forall j=0,1,\cdots$,
$\SAT_j$ is a sequence of elements in $\SAT$ such that every element is independent of its predecessors 
with respect to $\Qspace$. 
Hence, from the definition of eluder dimension, we have $|\SAT_j| \le \dimM{\Qspace}$, $\forall j=0,1,\cdots$.
Combining this result with Eqn(\ref{theorem1:1}), we have
$\sum_{j=0}^{J} \indicator \left[ R^{(j)}< V^*_0 (x_{j,0}) \right] \le |\SAT_{J+1}| \le \dimM{\Qspace}$, $\forall J=0,1,\cdots$. 
Finally, notice that $\sum_{j=0}^{J} \indicator \left[ V_j< V^*_0 (x_{j,0}) \right] $ is a non-decreasing function of $J$, and is bounded above by $\dimM{\Qspace}$.
Thus, 
$\lim_{J \rightarrow \infty} \sum_{j=0}^{J} \indicator \left[ R^{(j)}< V^*_0 (x_{j,0}) \right] = \sum_{j=0}^{\infty} \indicator \left[ R^{(j)} < V^*_0 (x_{j,0}) \right] $
exists, and satisfies
$\sum_{j=0}^{\infty} \indicator \left[ R^{(j)} < V^*_0 (x_{j,0}) \right] \le \dimM{\Qspace}$.
Hence we have $\left| \left \{j:\, R^{(j)} < V^*_0 (x_{j,0}) \right \} \right| \le \dimM{\Qspace}$. \textbf{q.e.d.}
\endproof

\section{Detailed Proof for Theorem \ref{the:state_aggregation} and Proposition \ref{the:performance_loss}}

\subsection{Proof for Lemma \ref{lemma:aggregation1}}
\proof{Proof for Lemma \ref{lemma:aggregation1}}
We prove Lemma \ref{lemma:aggregation1} by induction on $j$.
Note that when $j=0$, $\forall (x,a,t)$, $\Qopt_{j,t}(x,a)= \infty$.
Thus, Lemma \ref{lemma:aggregation1} trivially holds for $j=0$.

We now prove that if Lemma \ref{lemma:aggregation1} holds for episode
$j$, then it also holds for episode $j+1$, for any $j=0,1,\cdots$.
To prove this result, it is sufficient to show that
for any $(x,a,t)$ whose associated optimistic Q-value has been updated in episode $j$ (i.e.
$\Qopt_{j, t}(x,a) \neq \Qopt_{j+1, t}(x,a)$),  if the new optimistic Q-value 
$\Qopt_{j+1, t}(x,a)$ is still finite, then we have
$ |\Qopt_{j+1,t}(x,a)-Q^*_t(x,a)| \le 2 \rho (H-t)$. 

Note that if $\Qopt_{j, t}(x,a) \neq \Qopt_{j+1, t}(x,a)$, then $(x,a,t)$ must be in the same partition $\SAT_{t,k}$ as
$(x_{j,t}, a_{j,t}, t)$. Noting that $\sup_{Q \in \Qspace_{\CC_j}} \sup_{b \in \action} Q_{t+1} (x_{j,t+1}, b)  =\sup_{b \in \action} \Qopt_{j, t+1} (x_{j,t+1}, b)$,
from the discussion in Section \ref{sec:agnostic_learning}, we have
\[
\Qopt_{j+1, t}(x,a) = \overline{\theta}^{(j+1)}_{t,k}=
\left \{
\begin{array}{ll}
R_{H-1} (x_{j, H-1}, a_{j, H-1}) & \textrm{if $t=H-1$} \\
R_t (x_{j,t}, a_{j,t})+ \sup_{b \in \action} \Qopt_{j, t+1} (x_{j,t+1}, b) & \textrm{if $t<H-1$}
\end{array}
\right.
\]
We now prove $|\Qopt_{j+1, t}(x,a) -Q^*_t (x,a)| \leq 2 \rho (H-t)$ by considering two different scenarios.
First, if $t=H-1$, then $\Qopt_{j+1, t}(x,a)=R_{H-1} (x_{j, H-1}, a_{j, H-1})=Q^*_{H-1} (x_{j, H-1}, a_{j, H-1})$.  From our discussion above, we have
$|Q^*_t (x,a)- Q^*_{H-1} (x_{j, H-1}, a_{j, H-1})| \leq 2 \rho$, which implies that
$|Q^*_t (x,a)- \Qopt_{j+1, t}(x,a) | \leq 2 \rho = 2 \rho (H-t)$.
On the other hand,
if $t<H-1$, then
$\Qopt_{j+1, t}(x,a) = R_t (x_{j,t}, a_{j,t})+ \sup_{b \in \action} \Qopt_{j, t+1} (x_{j,t+1}, b)$.
If $\Qopt_{j+1, t}(x,a) < \infty$, then 
$\Qopt_{j, t+1} (x_{j,t+1}, b) < \infty$, $\forall b \in \action$.
Furthermore, from the induction hypothesis, $\Qopt_{j, t+1} (x_{j,t+1}, b) < \infty$, $\forall b \in \action$, implies that
$ \forall bÊ\in \action$, $\left| \Qopt_{j, t+1} (x_{j,t+1}, b)- Q^*_{t+1} (x_{j,t+1}, b) \right| \leq 2 \rho (H-t-1)$.
On the other hand, from the Bellman equation at $(x_{j,t}, a_{j,t}, t)$, we have that
$ Q^*_t (x_{j,t}, a_{j,t})  = R_t (x_{j,t}, a_{j,t}) + \sup_{b \in \action} Q^*_{t+1} (x_{j,t+1}, b) $. Thus,
\BE
\left| \Qopt_{j+1, t}(x,a)- Q^*_t (x_{j,t}, a_{j,t}) \right| &=& \left|  \sup_{b \in \action} \Qopt_{j, t+1} (x_{j,t+1}, b) - \sup_{b \in \action} Q^*_{t+1} (x_{j,t+1}, b) \right|  \nn
& \leq & \sup_{b \in \action} \left| \Qopt_{j, t+1} (x_{j,t+1}, b) - Q^*_{t+1} (x_{j,t+1}, b) \right|  \leq 2 \rho (H-t-1).  \nonumber
\EE
Moreover, since $(x,a,t)$ and $(x_{j,t}, a_{j,t}, t)$ are in the same partition, we have
$ \left| Q^*_t (x,a)- Q^*_t (x_{j,t}, a_{j,t}) \right| \leq 2 \rho$,
consequently, we have
$ \left| \Qopt_{j+1, t}(x,a)- Q^*_t (x, a) \right| \leq 2 \rho (H-t)$.
Thus, Lemma \ref{lemma:aggregation1} holds for episode $j+1$. 
By induction, we have proved Lemma \ref{lemma:aggregation1}. \textbf{q.e.d.}
\endproof
\noindent

\subsection{Proof for Lemma \ref{lemma:aggregation2}}
\proof{Proof for Lemma \ref{lemma:aggregation2}}
Notice that from OCP algoriothm, $\forall t=0,1,\cdots, H-1$, we have
$
\Qopt_{j, t} (x_{j,t}, a_{j,t}) \ge \Qopt_{j, t} (x_{j,t}, a)$, $\forall
a \in \action$.
Thus, if $\Qopt_{j,t} (x_{j,t}, a_{j,t}) < \infty$ for any $t$, then
$\Qopt_{j, t} (x_{j,t}, a) < \infty$, $\forall (a,t)$.
Consequently, from Lemma \ref{lemma:aggregation1}, we have that $\forall (a,t)$,
$\left | Q^*_t (x_{j,t} , a)- \Qopt_{j, t} (x_{j,t}, a) \right | \le 2 \rho (H-t)$.
Thus, for any $t$, we have
\[
Q^*_t (x_{j,t}, a_{j,t})+ 2 \rho (H-t) \ge
\Qopt_{j, t} (x_{j,t}, a_{j,t}) \ge \Qopt_{j, t} (x_{j,t}, a)
\ge Q^*_t (x_{j,t}, a) - 2 \rho (H-t), \quad \forall
a \in \action,
\]
which implies that $Q^*_t (x_{j,t}, a_{j,t}) \ge \sup_{a \in \action} Q^*_t (x_{j,t}, a) - 4 \rho (H-t)
= V^*_t (x_{j,t})- 4 \rho (H-t)$, $\forall t$.

We first prove that $V^*_{0}(x_{j,0})- R^{(j)} \le 2 \rho H(H+1)$.
Note that combining the above inequality with Bellman equation, we have that
$R_t (x_{j,t}, a_{j,t}) \ge V^*_t (x_{j,t})-V^*_{t+1}(x_{j,t+1})- 4 \rho (H-t)$ for any $t<H-1$
and $R_{H-1} (x_{j,H-1}, a_{j,H-1}) \ge V^*_{H-1} (x_{j,H-1})- 4 \rho$.
Summing up these inequalities, we have $V^*_0 (x_{j,0}) - R^{(j)} \leq 2 \rho H (H+1)$.

We now prove that $V^*_{0}(x_{j,0})- R^{(j)} \le 6 \rho H$ if the conditions of Proposition \ref{the:performance_loss} hold.
Note that the conditions of Proposition \ref{the:performance_loss} imply that
$U_{j,t} \geq \Qopt_{j,t} (x_{j,t}, a_{j,t}) \geq \Qpes_{j,t} (x_{j,t}, a_{j,t}) \geq L_{j,t}$ for any $t$. 
Note that by definition, 
$U_{j,H-1}=L_{j,H-1}=R_{H-1}(x_{j, H-1}, a_{j, H-1})$, and for $t<H-1$, we have
$
U_{j,t}=
R_t (x_{j,t}, a_{j,t})+  \Qopt_{j,t+1} (x_{j,t+1}, a_{j,t+1})$,
and
\BE
\vspace{-1cm}
L_{j,t}  
 \geq  R_t (x_{j,t}, a_{j,t}) + \sup_{a \in \action} \Qpes_{j, t+1} (x_{j,t+1}, a) 
\geq R_t (x_{j,t}, a_{j,t})  +  \Qpes_{j, t+1} (x_{j,t+1}, a_{j,t+1}),  \nonumber
\vspace{-1cm}
\EE
where the first inequality follows from the definition of $L_{j,t}$ and max-min inequality, and the second inequality follows from the fact that
$a_{j,t+1} \in \action$.
Combining the above inequalities, we have 
$\Qpes_{j,0} (x_{j,0}, a_{j,0}) \geq \sum_{t=0}^{H-1} R_t (x_{j,t}, a_{j,t})= R^{(j)} \geq \Qopt_{j,0} (x_{j,0}, a_{j,0}) \geq \Qpes_{j,0} (x_{j,0}, a_{j,0})$.
Thus we have $\Qopt_{j,0} (x_{j,0}, a_{j,0})=\Qpes_{j,0} (x_{j,0}, a_{j,0})=R^{(j)} < \infty$.
So from Lemma \ref{lemma:aggregation1},
$
\left| R^{(j)}- Q^*_0 (x_{j,0}, a_{j,0}) \right|= \left| \Qopt_{j,0} (x_{j,0}, a_{j,0}) - Q^*_0 (x_{j,0}, a_{j,0}) \right| \leq 2 \rho H$.
Thus, $R^{(j)} \geq Q^*_0 (x_{j,0}, a_{j,0}) - 2 \rho H$.
Furthermore, from the above analysis,
$Q^*_0 (x_{j,0}, a_{j,0}) \geq V_0^* (x_{j,0}) - 4 \rho H $. Thus we have
$R^{(j)} \geq V_0^* (x_{j,0}) - 6 \rho H $. \textbf{q.e.d.}
\endproof
\noindent

\subsection{Proof for Lemma \ref{state_aggregation:lemma3}}
\proof{Proof for Lemma \ref{state_aggregation:lemma3}}
$\forall j=0,1,\cdots$, if $t_j^* =\NULL$, then by definition of $t_j^*$ and Lemma \ref{lemma:aggregation2}, we have
$V^*_{0}(x_{j,0})- R^{(j)} \le 2 \rho H(H+1)$. On the other hand, if $t_j^* \neq \NULL$, then by definition of $t_j^*$, $\Qopt_{j,t_j^*} (x_{j,t_j^*}, a_{j,t_j^*}) =\infty$. 
We now show that $\Qopt_{j',t_j^*} (x_{j,t_j^*}, a_{j,t_j^*}) < \infty$  for all $j' > j$, 
and $\Qopt_{j',t_j^*} (x_{j,t_j^*}, a_{j,t_j^*}) = \infty$ for all $j' \leq j$.

Assume that $(x_{j,t_j^*}, a_{j,t_j^*}, t_j^*)$ belongs to partition $\SAT_{t_j^*,k}$, thus
$\Qopt_{j',t_j^*} (x_{j,t_j^*}, a_{j,t_j^*}) = \overline{\theta}^{(j')}_{t_j^*,k}$,
$\forall j'$.
Based on our discussion above, $\overline{\theta}^{(j')}_{t_j^*,k}$
is monotonically non-increasing in $j'$. Thus, $\Qopt_{j',t_j^*} (x_{j,t_j^*}, a_{j,t_j^*})$ is monotonically
non-increasing in $j'$, and hence for any $j' \leq j$, we have 
$\Qopt_{j',t_j^*} (x_{j,t_j^*}, a_{j,t_j^*}) = \infty$.
Furthermore, to prove that $\Qopt_{j',t_j^*} (x_{j,t_j^*}, a_{j,t_j^*}) < \infty$ for all $j' > j$,
it is sufficient to prove that
$ \Qopt_{j+1,t_j^*} (x_{j,t_j^*}, a_{j,t_j^*}) < \infty $.

From OCP, the algorithm will add a new constraint
$L_{j,t_j^*} \leq Q_{t_j^*} (x_{j,t_j^*}, a_{j,t_j^*}) \leq U_{j,t_j^*}$.
We first prove that 
$U_{j,t_j^*} < \infty$.
To see it, notice that if $t_j^* =H-1$, then
$U_{j,t_j^*}=U_{j,H-1}=R_{H-1} (x_{j,H-1}, a_{j, H-1}) < \infty$.
On the other hand, if $t_j^* < H-1$, then by definition
$U_{j,t_j^*} =
R_{t_j^*} (x_{j,t_j^*}, a_{j, t_j^*}) +  \Qopt_{j,t_j^*+1} (x_{j,t_j^*+1}, a_{j,t_j^*+1})$.
From the definition of $t_j^*$, $\Qopt_{j,t_j^*+1} (x_{j,t_j^*+1}, a_{j,t_j^*+1}) < \infty$, thus
$U_{j,t_j^*} < \infty$. Consequently,
$
\Qopt_{j+1,t_j^*} (x_{j,t_j^*}, a_{j,t_j^*}) = \overline{\theta}^{(j+1)}_{t_j^*,k} = \min \{  \overline{\theta}^{(j)}_{t_j^*,k}, U_{j,t_j^*}  \} \leq U_{j,t_j^*} < \infty
$.
Thus,  
$\Qopt_{j',t_j^*} (x_{j,t_j^*}, a_{j,t_j^*}) < \infty$ for all $j' >j$. 

Thus, if we consider $\Qopt_{j', t_j^*} (x_{j,t_j^*}, a_{j,t_j^*}) = \overline{\theta}^{(j')}_{t_j^*,k}$ as a function of $j'$,
then this function transits from infinity to finite values in episode $j$.
In summary, $t_j^* \neq \NULL$ implies that $\overline{\theta}^{(j')}_{t_j^*,k}$
transits from infinity to finite values in episode $j$.
Since other $\overline{\theta}^{(j')}_{t,k}$'s might also transit from $\infty$ to finite values in episode $j$, thus
$\indicator [t_j^* \neq \NULL]$ is less than or equal to the number of $\overline{\theta}^{(j')}_{t,k}$'s transiting from $\infty$ to finite values in
episode $j$.
Note that from the monotonicity of $\overline{\theta}^{(j')}_{t,k}$, for each partition, this transition can occur at most once,
and there are $K$ partitions in total. Hence we have
$ \sum_{j=0}^{\infty} \indicator [t_j^* \neq \NULL] \leq K$. \textbf{q.e.d.}
\endproof
\noindent

\end{APPENDICES}

\bibliographystyle{ormsv080} 
\bibliography{reference} 

\end{document}